\pdfoutput=1
\documentclass[11pt]{article}

\usepackage[final]{acl}
\usepackage{times}
\usepackage{latexsym}
\usepackage[T1]{fontenc}
\usepackage[utf8]{inputenc}

\usepackage{microtype}
\usepackage{inconsolata}
\usepackage{graphicx}
\usepackage{subcaption}
\usepackage{booktabs}
\usepackage{array}
\usepackage{ragged2e}
\usepackage{xurl}
\usepackage{xcolor}

\newcommand{\theHalgorithm}{\arabic{algorithm}}

\usepackage{amsmath}
\usepackage{amssymb}
\usepackage{mathtools}
\usepackage{amsthm}
\usepackage[capitalize,noabbrev]{cleveref}
\usepackage{float}
\usepackage[section]{placeins}
\usepackage{stfloats}
\usepackage{algorithm}
\usepackage{algorithmic}
\usepackage{fancyvrb}
\usepackage{fvextra}

\DefineVerbatimEnvironment{PromptBox}{Verbatim}{fontsize=\scriptsize,frame=single,framesep=1mm,breaklines=true,breakanywhere=true,breaksymbolleft={},breaksymbolright={}}

\newcommand{\CompOK}{\mathsf{Comp}}
\newcommand{\SemOK}{\mathsf{Sem}}

\title{FormalEvolve: Neuro-Symbolic Evolutionary Search for Diverse Autoformalization}
\author{
  Haijian Lu$^{1,2}$ \quad Wei Wang$^{2}$\thanks{Corresponding author: \texttt{wangwei@nlpr.ia.ac.cn}.} \quad Jing Liu$^{1}$\\
  $^{1}$School of Artificial Intelligence, Xidian University, Xi'an, China\\
  $^{2}$State Key Laboratory of General Artificial Intelligence, BIGAI, Beijing, China
}

\hypersetup{
  pdfauthor={Haijian Lu, Wei Wang, Jing Liu},
  pdftitle={FormalEvolve: Neuro-Symbolic Evolutionary Search for Diverse Autoformalization},
  pdfsubject={},
  pdfkeywords={Autoformalization, Lean 4, Evolutionary Search, Formal Mathematics}
}

\newcommand{\PUCBPassOurs}{44/100}
\newcommand{\PUCBCompleteOurs}{27/100}
\newcommand{\PUCBTheoremCompleteOurs}{13/100}
\newcommand{\PUCBPassSample}{41/100}
\newcommand{\PUCBCompleteSample}{23/100}
\newcommand{\PUCBTheoremCompleteSample}{8/100}
\newcommand{\PUCBPassStrongSemantic}{40/100}
\newcommand{\PUCBCompleteStrongSemantic}{23/100}
\newcommand{\PUCBTheoremCompleteStrongSemantic}{8/100}
\newcommand{\PUCBPassHybrid}{40/100}
\newcommand{\PUCBCompleteHybrid}{22/100}
\newcommand{\PUCBTheoremCompleteHybrid}{9/100}

\newcommand{\PUCBPassGT}{96/100}
\newcommand{\PUCBCompleteGT}{68/100}
\newcommand{\PUCBTheoremCompleteGT}{18/100}

\newcommand{\PUPNPassOurs}{127/186}
\newcommand{\PUPNCompleteOurs}{52/186}
\newcommand{\PUPNTheoremCompleteOurs}{45/186}
\newcommand{\PUPNPassSample}{106/186}
\newcommand{\PUPNCompleteSample}{46/186}
\newcommand{\PUPNTheoremCompleteSample}{41/186}
\newcommand{\PUPNPassStrongSemantic}{119/186}
\newcommand{\PUPNCompleteStrongSemantic}{50/186}
\newcommand{\PUPNTheoremCompleteStrongSemantic}{46/186}
\newcommand{\PUPNPassHybrid}{110/186}
\newcommand{\PUPNCompleteHybrid}{49/186}
\newcommand{\PUPNTheoremCompleteHybrid}{40/186}

\newcommand{\PUPNPassGT}{146/186}
\newcommand{\PUPNCompleteGT}{46/186}
\newcommand{\PUPNTheoremCompleteGT}{34/186}

\begin{document}
\raggedbottom

\maketitle

\begin{abstract}
Autoformalization aims to produce formal statements that compile and faithfully preserve the intended meaning of informal mathematics.
Yet standard single-output evaluation collapses this many-to-many structure into a single prediction.
For downstream proving, this granularity is too coarse: a formal statement is not merely a faithful translation endpoint, but also a prover-facing interface whose structure can alter proof search under a fixed budget.
We therefore recast autoformalization as budgeted test-time search: FormalEvolve maintains a compilation-feasible archive for reuse and returns a deduplicated, semantically accepted repertoire for evaluation and downstream proving.
It expands the archive with LLM-driven mutation, crossover, bounded patch repair, and symbolic abstract syntax tree (AST) rewrites for structural diversity.
Under a generator-call budget of $T{=}100$ with a fixed LLM semantic judge, FormalEvolve reaches SH@100 of 58.0\% on CombiBench and 84.9\% on ProofNet, improving over all no-archive controls while reducing the cross-problem concentration of semantic successes.
Under a fixed $B{=}64$ prover budget, these repertoires improve theorem-complete proving over the matched no-archive control.
Additional stronger-base statement-generation experiments show that archive-search gains persist with stronger seed and repair models.
\end{abstract}

\section{Introduction}
\label{sec:intro}

\begin{figure}[t]
  \centering
  \includegraphics[width=\columnwidth]{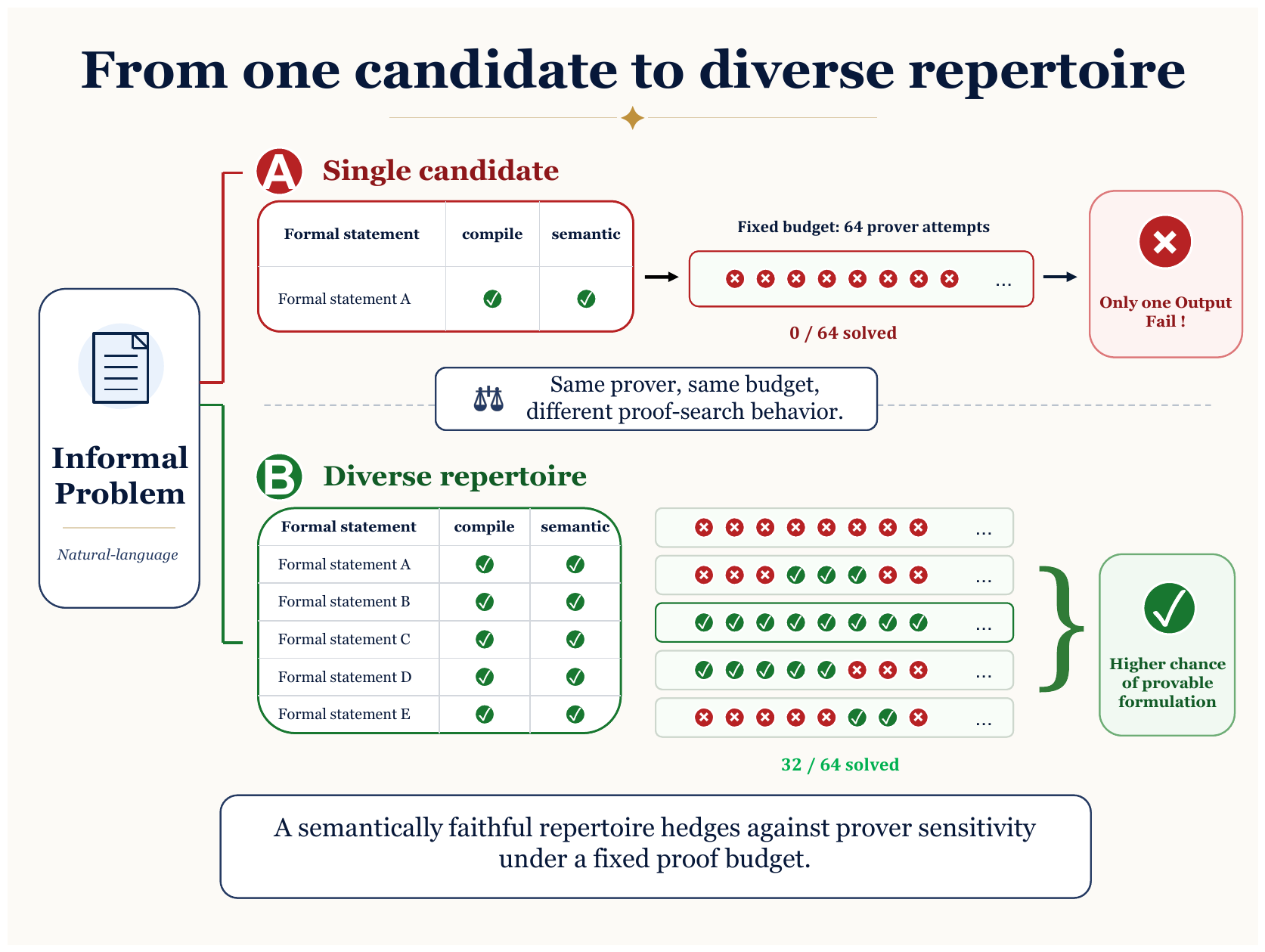}
  \caption{Motivating illustration under a fixed prover and attempt budget ($B{=}64$). Even when candidate statements compile and are judged semantically consistent, prover outcomes can vary substantially. Constructing a diverse repertoire hedges against this sensitivity and increases the chance that at least one provable statement is found within budget.}
  \label{fig:teaser_statement_sensitivity}
\end{figure}

\begin{figure*}[t]
  \centering
  \includegraphics[width=0.94\textwidth]{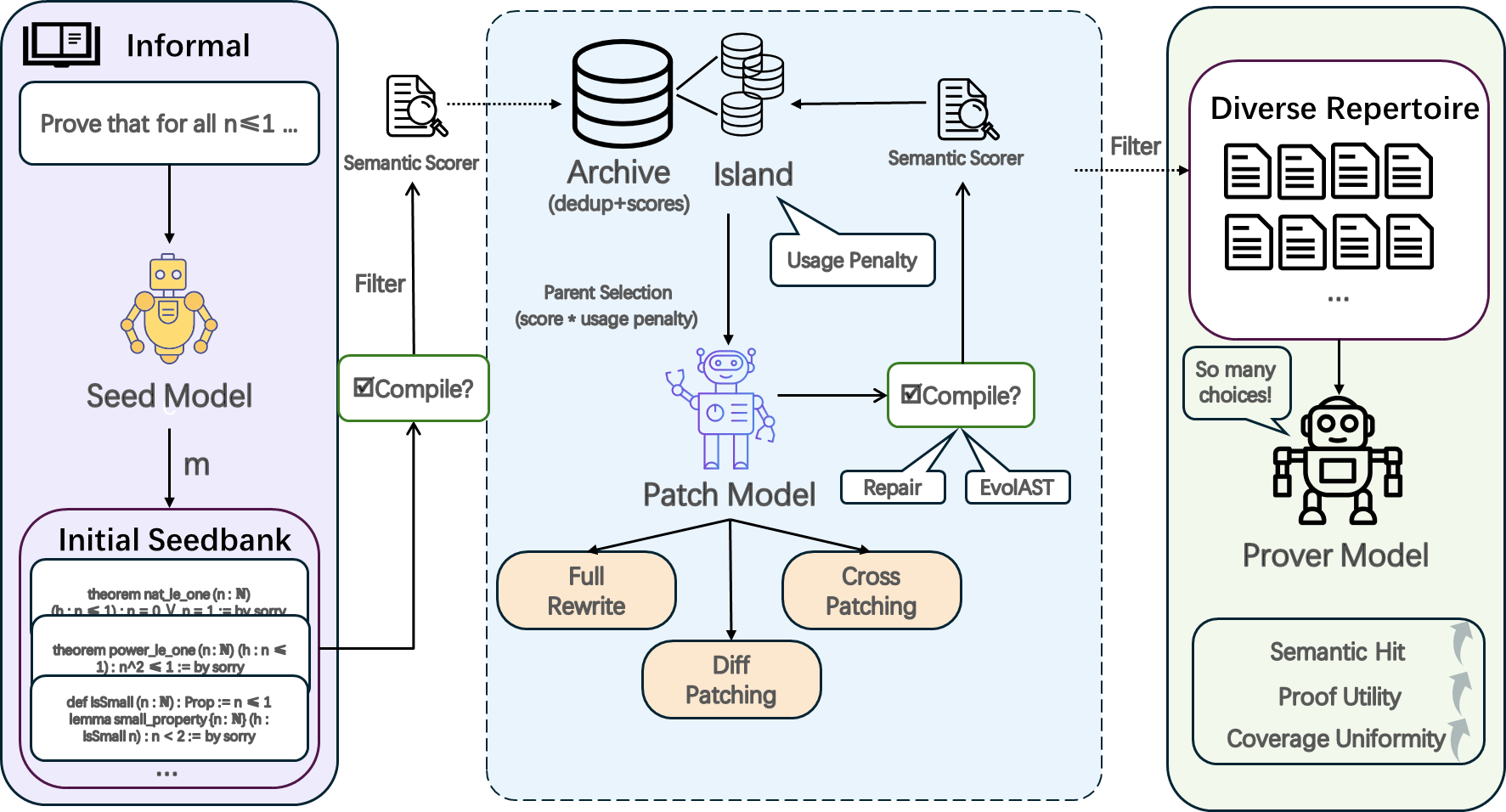}
  \caption{Framework overview of FormalEvolve. A seed model produces an initial seedbank (debited as generator calls) that initializes a compilation-feasible archive. Candidates are stored with semantic scores and selected with usage penalties; a patch model proposes edits (full rewrite, diff patching, and cross patching) and triggers bounded repair or EvolAST fallbacks on duplicate proposals and compilation failures. Semantic scoring filters the archive into a diverse repertoire for downstream proving and evaluation.}
  \label{fig:framework}
\end{figure*}

Autoformalization translates informal mathematics into Lean statements that compile and preserve the intended mathematical claim.
For downstream proving, however, the formal statement is not merely a translation endpoint: it is the prover-facing interface through which proof search is attempted.
Compilation requires successful elaboration and type checking under the chosen context, while semantic faithfulness remains a separate requirement.
As large language model provers become more capable \citep{polu2020generative,han2021proofartifact,song2024leancopilot}, structural choices in the statement itself increasingly affect downstream proof search.

Autoformalization is naturally many-to-many: the same theorem can often be encoded as multiple faithful Lean statements.
Semantic consistency therefore does not imply prover equivalence: by \emph{prover equivalence} we informally mean comparable proof-search behavior under the same prover and attempt budget.
A formal statement that compiles and preserves the intended claim is not necessarily a prover-friendly target; faithful variants can differ substantially in proof-search cost and success rate.
\Cref{fig:teaser_statement_sensitivity} illustrates this gap: candidates that pass compilation and semantic checks can still lead to different prover outcomes under the same prover protocol.
A single generated statement therefore discards prover-relevant variation.
This motivates a repertoire view: the search procedure should construct multiple faithful Lean statements that expose different proof-search behavior under the same prover budget.

We formulate autoformalization as a budgeted test-time search problem.
Under a generator-call budget, the system must construct a repertoire of candidate formalizations that are compilable, semantically consistent, and useful for downstream proving.
This requires preserving compilation feasibility and semantic fidelity while distributing search across distinct trajectories and avoiding near-duplicate candidates.

To address these requirements, we propose FormalEvolve, a compilation-gated neuro-symbolic evolutionary framework built around an archive search protocol.
FormalEvolve maintains a compilation-feasible archive of candidate Lean statements and expands it primarily through LLM-driven mutation, crossover, and bounded patch repair.
In addition, conservative generator-call-free AST rewrites act as a structural diversification fallback when patching stalls, for example after repeated duplicates or compile failures.
Usage penalties and duplicate-aware selection encourage the search to cover more problems and statement forms.
\Cref{fig:framework} gives a high-level overview.

Under a generator-call budget of $T{=}100$, FormalEvolve reaches SH@100 of 58.0\% on CombiBench and 84.9\% on ProofNet, improving over all no-archive controls on both benchmarks while reducing the cross-problem concentration of semantic successes.
Under a fixed $B{=}64$ prover budget, the resulting repertoires improve theorem-complete downstream proving over the matched no-archive control on both benchmarks.
To separate repair-model strength from archive search, we include a matched Hybrid control that uses the same stronger repair model without an archive; FormalEvolve still gains at the statement stage. The pattern also holds in a stronger-base statement-generation setting.
Because semantic labels come from an LLM judge, we also provide manual audits at both the statement and theorem-complete stages to calibrate the evaluation.

More broadly, FormalEvolve illustrates a verifier-gated diversity-search pattern: when many valid formal artifacts can express the same informal intent, test-time compute can construct a repertoire of useful alternatives under hard checker feedback.

We make three primary contributions:
\begin{enumerate}
    \item \textbf{Problem Modeling and Protocol:} We formulate Lean~4 autoformalization as budgeted test-time repertoire search and introduce budget-auditable metrics that connect statement-level coverage, cross-problem concentration, and downstream proving performance.
    \item \textbf{Search Mechanism:} We propose FormalEvolve, a compilation-gated neuro-symbolic evolutionary framework that combines archive-based LLM mutation and crossover, bounded patch repair, and generator-call-free symbolic AST rewrites.
    \item \textbf{Empirical Evidence:} We show that FormalEvolve improves semantic hit rates and cross-problem coverage uniformity under fixed generator-call budgets, improves theorem-complete downstream proving over a matched no-archive control by constructing diverse candidate repertoires, and preserves statement-stage archive-search gains under a stronger-base setting.
\end{enumerate}

\section{Related Work}
\label{sec:related}

\paragraph{Autoformalization and semantic evaluation.}
Large language models have enabled autoformalization at scale, with systematic evaluation on curated formal-mathematics benchmarks \citep{wu2022autoformalization,azerbayev2023proofnet,zheng2021minif2f}.
Reliability remains a central bottleneck: compilation checks syntax and elaboration, but is only a weak proxy for semantic consistency, motivating dedicated semantic judges and alignment evaluators \citep{peng2025criticlean,lu2024formalalign}.
Recent systems further improve semantic fidelity by interleaving iteration, tool feedback, and grounding \citep{reform2025,guo2025atf,wang2025aria}.
These lines of work primarily improve or assess individual formalizations within a single-output view.
We study fixed-budget construction of semantically consistent and diverse candidate repertoires.

\paragraph{Statement variation and prover utility.}
For downstream proving, a formal statement serves both as a semantic target and as the object consumed by a prover.
Even among semantically consistent candidates, minor reformulations intended to preserve semantics can substantially change proof-search difficulty \citep{zhao2025ineqcomp}.
At the same time, semantic evaluation itself remains imperfect \citep{reform2025}.
These observations motivate treating autoformalization as a fixed-budget repertoire-construction problem, where multiple faithful candidates expose different prover utilities.
Such a repertoire mitigates prover sensitivity by exposing several semantically aligned targets under the same fixed budget.

\paragraph{Test-time search and quality-diversity.}
Test-time search and evolutionary optimization with LLMs have been explored for sample-efficient program and algorithm improvement \citep{novikov2025alphaevolve,shinkaevolve2026,liu2024llm4ad}.
Quality-diversity methods construct repertoires of high-quality solutions \citep{lehman2011novelty,mouret2015mapelites}, and multi-criteria evolutionary optimization provides a complementary perspective on multiple objectives \citep{deb2002nsga2}.
EvolProver synthesizes offline prover-training data through symmetry, difficulty, and semantics-preserving AST rewrites \citep{tian2025evolprover}.
FormalEvolve performs online, per-problem test-time search and adapts conservative AST rewrites as a targeted diversification fallback.
For Lean statement search, elaboration and type checking define the feasible set: a candidate must compile before it can enter the archive or be reused as a parent.
For each informal problem, FormalEvolve builds a test-time repertoire of candidate statements used for that problem's evaluation and downstream proving.

\section{Method}
\label{sec:method}

FormalEvolve instantiates the shift from single-output prediction to many-to-many repertoire construction as per-problem test-time search over Lean~4 statement candidates under a generator-call budget $T$ (\Cref{sec:budget}).
Given an informal statement $x$, the goal is to construct a deduplicated repertoire of candidates that compile and pass the semantic judge within budget.
The search starts from a small compilation-feasible seedbank, stores unique compiled candidates in an archive, and expands it through LLM patching, bounded repair, and call-free EvolAST diversification.
Because Lean compilation defines a narrow feasible region, archive entry is compilation-gated, near-feasible candidates are salvaged by bounded repair, and lightweight diversity operators discourage repeated near-duplicate trajectories.
Our reported output is the deduplicated semantically consistent repertoire generated within budget, i.e., candidates satisfying $\CompOK(c){=}1$ and $\SemOK(c){=}1$, which we use for SH@T and downstream proving.

\Cref{alg:formalevolve} summarizes the same loop and makes the two domain gates explicit: $\CompOK$ checks whether a candidate passes Lean~4 compilation, while $\SemOK$ is the semantic-faithfulness judge against the informal theorem $x$.
The counter $t$ tracks debited generator calls, $\mathcal{A}_I$ is an island archive, $\mathcal{Z}$ denotes sampled prompt context and inspirations, $\mathcal{B}$ is the candidate batch produced by proposal and bounded repair, and $\mathcal{G}$ is the persistent accepted repertoire returned at the end.
\Cref{fig:framework} gives the visual overview, and \Cref{sec:appendix-pseudocode} gives the detailed pseudocode.

\subsection{Search Procedure and State}

\paragraph{Basic objects and gates.}
We search over Lean~4 candidates of the form $c=(\hat{h},y)$, where $\hat{h}$ is an import/header context and $y$ is the primary Lean~4 statement or declaration target.
Most candidates are theorem/lemma-style files with placeholder proofs, but benchmark contexts and model outputs can also contain \texttt{def}, \texttt{abbrev}, or \texttt{example} templates; for this reason, the prover-stage evaluation separately reports theorem-complete success for explicit \texttt{theorem}/\texttt{lemma} completions.
We evaluate candidates using a compilation predicate $\CompOK(c)\in\{0,1\}$ and a semantic predicate $\SemOK(c)\in\{0,1\}$ (instantiated by an LLM judge in \Cref{sec:exp}).
Compilation is a hard feasibility gate: only candidates with $\CompOK(c)=1$ are eligible to enter the archive and be reused as parents/inspirations.
We treat compilation/elaboration as verifier feedback: it both defines feasibility and provides the primary error feedback for bounded compilation repair.
During search, the archive stores unique compilation-feasible candidates regardless of $\SemOK(c)$ (i.e., it may include semantically inconsistent stepping stones) and is used only for parent/context sampling.
For reporting SH@T and downstream proving, we define the \emph{semantically consistent repertoire} as the deduplicated set of candidates generated within budget that satisfy both $\CompOK(c)=1$ and $\SemOK(c)=1$.
FormalEvolve uses these two predicates during test-time search and bounded repairs; downstream proving is evaluated only when reporting proof utility.
All generator-side LLM calls (seeds, patches, and repairs) are debited against the same per-problem generator-call budget $T$ (\Cref{sec:budget}).

\paragraph{Seedbank initialization.}
We initialize the search by sampling a small seedbank with a seed model $M_{\mathrm{seed}}$, evaluating each seed with the compiler and semantic judge, and inserting the compilation-feasible subset into the archive.
When a pre-sampled seedbank is reused across methods, we still debit each reused seed as one generator call to preserve consistent accounting.
We separate the seed and patch roles: $M_{\mathrm{seed}}$ provides initial candidates, while $M_{\mathrm{patch}}$ performs edit-conditioned proposals and bounded repairs.
Concrete model choices are specified in \Cref{sec:exp}.

\paragraph{Archive, islands, and duplicates.}
We maintain a per-problem archive $\mathcal{A}$ of unique compilation-feasible candidates that serves both as a parent pool and as prompt context.
We use an island model with $K$ semi-isolated archive populations (default $K=2$).
When $K=1$, FormalEvolve uses a single archive population: it still retains archive-based parent reuse, archive-conditioned context sampling, usage-penalized selection, and bounded repair, but removes island partitioning and migration.
Each iteration samples one island and restricts parent/context sampling to it.
In the main configuration, migration occurs every 10 generations at rate 0.1 with the top candidate in each island protected.
Detailed hyperparameters are provided in \Cref{sec:appendix-config}.
Exact duplicates under canonicalization (whitespace/name normalization) are rejected at archive insertion to reduce collapse.
In \Cref{alg:formalevolve}, \textsc{SampleParent} uses the scoring rule defined below, and $n_i$ tracks how often $c_i$ has already been selected as a parent.

\begin{algorithm}[t]
  \caption{FormalEvolve for Lean statement search}
  \label{alg:formalevolve}
  \footnotesize
  \begin{algorithmic}
    \STATE \textbf{Task:} given informal theorem $x$, search Lean~4 statement candidates $c$
    \STATE \textbf{Gates:} $\CompOK(c)$ checks whether $c$ passes Lean~4 compilation; $\SemOK(c)$ checks faithfulness to $x$
    \STATE \textbf{Input:} budget $T$, models $M_{\mathrm{seed}},M_{\mathrm{patch}}$, islands $K$
    \STATE \textbf{Output:} accepted repertoire $\mathcal{G}$
    \STATE $t\leftarrow 0$; $\mathcal{A}\leftarrow$ empty $K$-island archive; $\mathcal{G}\leftarrow\emptyset$
    \STATE $(\mathcal{A},t)\leftarrow\textsc{SeedArchive}(x,M_{\mathrm{seed}},T,K)$ \hfill \textit{(Seedbank initialization)}
    \STATE $\mathcal{G}\leftarrow\{c\in\bigcup_I\mathcal{A}_I:\CompOK(c)\land\SemOK(c)\}$
    \WHILE{$t<T$ and $\bigcup_I\mathcal{A}_I\neq\emptyset$}
      \STATE $I\leftarrow\textsc{SampleIsland}(\mathcal{A})$ \hfill \textit{(Archive, islands, and duplicates)}
      \STATE $p\leftarrow\textsc{SampleParent}(\mathcal{A}_I)$; $n_p\leftarrow n_p+1$ \hfill \textit{(Scoring; Eq.~\ref{eq:parent-weight})}
      \STATE $\mathcal{Z}\leftarrow\textsc{SampleContext}(\mathcal{A}_I,p)$
      \STATE $c\leftarrow\textsc{Propose}(p,\mathcal{Z},x;M_{\mathrm{patch}})$; $t\leftarrow t+1$ \hfill \textit{(Variation operators)}
      \STATE $(\mathcal{B},t)\leftarrow\textsc{ExpandAndRepair}(c,p,x,t,T)$ \hfill \textit{(Bounded repair + EvolAST fallback)}
      \STATE $\mathcal{A}_I\leftarrow\textsc{EvaluateAndUpdate}(\mathcal{A}_I,\mathcal{B})$
      \STATE $\mathcal{G}\leftarrow\mathcal{G}\cup\{b\in\mathcal{B}:\CompOK(b)\land\SemOK(b)\}$
    \ENDWHILE
    \STATE \textbf{return} $\mathrm{Dedup}(\mathcal{G})$
  \end{algorithmic}
\end{algorithm}

\paragraph{Scoring and usage-penalized selection.}
The selection rule separates archive storage from parent sampling.
Storage is purely feasibility-gated: only candidates with $\CompOK(c)=1$ can enter the archive, and the archive retains all unique compilation-feasible candidates.
This matters because different faithful statements can expose different binder structures, imports, and theorem shapes to the downstream prover.
Scores therefore bias reuse as future parents; archive contents remain available for later sampling, and proof utility is evaluated downstream under the fixed prover protocol.

For an archived candidate $c_i\in\mathcal{A}_I$, let $n_i$ be how many times it has already been selected as a parent.
We assign each candidate a sampling weight $w_i$ and sample proportionally:
\begin{equation}
  \Pr(p=c_i\mid\mathcal{A}_I)
  =
  \frac{w_i}{\sum_{c_j\in\mathcal{A}_I} w_j}.
  \label{eq:parent-sampling}
\end{equation}
The weight combines the gate score and a usage discount:
\begin{equation}
  \begin{aligned}
    s_i &= \CompOK(c_i)(1+\SemOK(c_i)),\\
    \tilde{s}_I &= \mathrm{median}_{c_j\in\mathcal{A}_I}(s_j),\\
    d_I &= \max(\mathrm{MAD}(\{s_j:c_j\in\mathcal{A}_I\}),\varepsilon),\\
    z_i &= (s_i-\tilde{s}_I)/d_I,\\
    u_i &= [1+(1+\beta)n_i]^{-1},\\
    w_i &= \sigma(\lambda z_i)u_i .
  \end{aligned}
  \label{eq:parent-weight}
\end{equation}
Since archived candidates already compile, $s_i=1$ for judge-negative archive members and $s_i=2$ for judge-consistent members.
$u_i$ is a usage discount: it equals 1 for a never-used parent and decreases as the same parent is sampled repeatedly.
We set $\varepsilon=10^{-6}$ only to avoid division by zero when all scores are identical, and use $\lambda=10$ and $\beta=0.05$ in the main runs.

\subsection{Proposal, Repair, and Diversification}

\paragraph{Variation operators.}
Given a selected parent, the patch model proposes a new candidate using one of three prompt templates.
\emph{Full} patching rewrites the entire statement (and, when allowed, its surrounding imports) conditioned on the informal input and parent context.
\emph{Diff} patching performs localized edits that make minimal changes relative to the parent, guided by compiler/judge feedback.
\emph{Cross} patching additionally conditions on one or more \emph{inspiration} candidates sampled from the current archive.
The model combines useful elements (e.g., import context, binder structure, or type annotations) while preserving the informal input's meaning (see \Cref{sec:appendix-prompts} for prompt details and \Cref{sec:appendix-case-studies} for audited examples).
All three templates return a complete Lean~4 file (imports plus a single statement) and are budgeted identically as generator calls.

\paragraph{Bounded compilation and semantic repair.}
We compile the full Lean~4 file and treat successful compilation as a hard feasibility gate.
When compilation fails, FormalEvolve invokes a repair prompt that proposes a minimal patch conditioned on compiler feedback.
Repair is bounded to a fixed cap of two repair attempts per proposal path in the main configuration; each attempt is debited as a generator call, and repair stops immediately when the remaining budget is exhausted.
Bounding repair makes the search budget-auditable under hard cutoffs and prevents pathological cases where a single hard candidate consumes most of the call budget.
For a compilable candidate that fails the semantic judge ($\CompOK(c)=1$ and $\SemOK(c)=0$), we optionally invoke bounded \emph{semantic repair}.
$M_{\mathrm{patch}}$ revises the statement using the informal input and the judge's rationale (when available), after which the result is re-evaluated; every repair and semantic-repair attempt is debited.
We view compilation repair and semantic repair as two instantiations of the same edit-conditioned patch mechanism, differing only in the feedback source (compiler errors vs.\ judge rationale).

\paragraph{Call-free diversity fallback (EvolAST).}
Archive-based search can suffer mode collapse, with many candidates becoming near-duplicates or exploiting superficial patterns.
FormalEvolve mitigates this primarily via the usage penalty in parent sampling.
We additionally apply a conservative, generator-call-free AST rewrite fallback inspired by EvolProver \citep{tian2025evolprover} when patching stalls (e.g., repeated duplicates or compile failures).
In the reported runs, each triggering proposal contributes at most one EvolAST representative, and EvolAST outputs are not recursively rewritten or sent into LLM repair.
EvolAST rewrites only within binder types and the goal type (imports/preamble unchanged) to produce symmetry/structure variants, which are then filtered by the same compilation gate and semantic judge.
EvolAST consumes no generator calls; \Cref{sec:appendix-budget-audit} reports its compilation and judge overhead.

\section{Experiments}
\label{sec:exp}
\label{sec:budget}

\paragraph{Benchmarks.}
We evaluate on ProofNet \citep{azerbayev2023proofnet} (Lean~4 port; test split, $N{=}186$) and CombiBench \citep{liu2025combibench} ($N{=}100$, domain/style shift).
The two benchmarks provide contrasting search regimes.
ProofNet consists primarily of standard undergraduate mathematics, and the public autoformalization SFT mixture used to initialize Kimina-Autoformalizer includes ProofNet \citep{wang2025kimina}, making it a high-prior setting for the Kimina-based generation component.
CombiBench instead focuses on specialized combinatorics and presents a substantially stronger domain/style shift, providing a lower-prior setting in which exploration is more important.

\paragraph{Environment.}
Candidates are compiled under a pinned Lean~4/Mathlib toolchain served by Kimina Lean Server \citep{dossantos2025kimina}; full details are in \Cref{sec:appendix-config}.

\paragraph{Budget.}
We use $T=100$ generator calls per problem.
Each seed, proposal, compilation-repair, and semantic-repair LLM call debits one call.
We additionally record compilation checks, semantic-judge calls, and EvolAST-triggered candidate evaluations; Appendix~\ref{sec:appendix-budget-audit} reports this accounting, including EvolAST-attributed judge calls.

\paragraph{Baselines.}
Our sampling baselines instantiate a memoryless, no-archive setting: budget-matched variants with fresh sampling/repair and no archive state.
They include \textbf{Sample} (no repair), \textbf{Compile Repair} (bounded compilation repair), and \textbf{C+S Repair} (Compile+Semantic Repair: compilation repair plus one bounded semantic-repair call for compilable candidates failing the judge).
To separate archive-search effects from patch-model strength, Table~\ref{tab:full-hit100} also reports \textbf{Hybrid control}: the same Kimina-generation/Qwen3-repair stack under $T{=}100$, but with independently sampled trajectories and no archive state, parent reuse, archive context, usage-penalized selection, or island migration.

\paragraph{Models.}
In our main runs, the sampling baselines use Kimina-Autoformalizer-7B for sampling/repair, while FormalEvolve uses Kimina-Autoformalizer-7B for seeding and Qwen3-30B-A3B for patch/repair.
For proof utility, we fix the prover to Goedel-Prover-V2-32B \citep{lin2025goedelproverv2} with a fixed prompt template (\Cref{sec:appendix-prover-prompt}).

\paragraph{Search configuration.}
Unless stated otherwise, FormalEvolve uses $K=2$ islands, archive capacity 40, parent-selection parameters $\lambda=10$ and $\beta=0.05$, an operator mix of full/diff/cross patching with probabilities 0.5/0.3/0.2, and migration every 10 generations at rate 0.1; Appendix~\ref{sec:appendix-config} gives the complete configuration.

\paragraph{Metrics.}
CH@T and SH@T denote the fraction of problems with at least one compilable candidate and at least one compilable, judge-consistent candidate, respectively, within $T$ debited generator calls.
Coverage alone leaves open whether additional semantic successes are broadly distributed or concentrated on a small set of already-easy problems.
Let $s_j(T)$ be the number of generated candidates with $\CompOK(c)=1$ and $\SemOK(c)=1$ for problem $j$ within budget $T$; we therefore report cross-problem concentration via the Gini coefficient
\begin{equation}
  \mathrm{Gini}(T)=\frac{\sum_{i=1}^{N}\sum_{j=1}^{N} |s_i(T)-s_j(T)|}{2N\sum_{j=1}^{N} s_j(T)+\varepsilon},
\end{equation}
with the same numerical stabilizer $\varepsilon=10^{-6}$ used in parent-score normalization.
We also report the Top-10\% share $\sum_{j\in \mathrm{Top}_{10\%}} s_j(T)/\sum_{j=1}^{N} s_j(T)$ (where $\mathrm{Top}_{10\%}$ selects the problems with the largest $s_j(T)$).
We evaluate downstream proof utility under a fixed round-robin $B=64$ prover-attempt protocol (pass@64 / complete@64 / theorem-complete@64; \Cref{tab:proof-utility-64}); full definitions are in \Cref{sec:appendix-metric-definitions}.

\paragraph{Semantic judge and manual calibration.}
The semantic predicate $\SemOK$ is implemented by CriticLean-Qwen3-14B with a fixed prompt and serving configuration; the judge prompt and judge-call accounting are reported in Appendix~\ref{sec:appendix-prompts-judge} and Appendix~\ref{sec:appendix-budget-audit}.
Our semantic metrics rely on an LLM-as-judge implementation of $\SemOK$.
We therefore include manual faithfulness audits for a statement-stage sample from judge-positive ProofNet outputs and a theorem-complete prover-stage sample produced under the fixed $B=64$ prover protocol.
We summarize both audits in \Cref{sec:results} and provide their protocols and tables in Appendix~\ref{sec:appendix-manual-faithfulness-audit}.

\section{Results}
\label{sec:results}

\subsection{Coverage and Uniformity}

\paragraph{Statement generation at $T=100$.}
Table~\ref{tab:full-hit100} reports statement generation under the same $T=100$ generator-call budget, including coverage (CH/SH) and concentration (Gini/top-10\% share).
Against the hybrid no-archive control, which uses Kimina generation with the same Qwen3 repair model but no archive, FormalEvolve increases SH@100 from 0.530 to 0.580 on CombiBench ($+0.050$, paired 95\% CI $[0.003, 0.130]$) and from 0.828 to 0.849 on ProofNet ($+0.022$, paired 95\% CI $[0.005, 0.048]$).
We therefore interpret coverage together with cross-problem concentration and downstream proof utility.

\begin{table*}[t]
			  \centering
\caption{Statement generation at $T=100$. CH/SH measure compile/semantic coverage; Gini and top-10\% share measure concentration of semantic-success counts (lower is better).}
		  \label{tab:full-hit100}
		  \scriptsize
\resizebox{\textwidth}{!}{%
		  \begin{tabular}{lcccc|cccc}
	  \toprule
	   & \multicolumn{4}{c}{ProofNet (test, $N=186$)} & \multicolumn{4}{c}{CombiBench ($N=100$)} \\
		  Method & CH@100 & SH@100 & Gini$\downarrow$ & Top-10\%$\downarrow$ & CH@100 & SH@100 & Gini$\downarrow$ & Top-10\%$\downarrow$ \\
		  \midrule
		  \multicolumn{9}{l}{\textbf{Baselines (Kimina; no archive)}} \\
		  Sample & 0.903 & 0.715 & 0.537 & 0.243 & 1.000 & 0.440 & 0.816 & 0.637 \\
		  Compile Repair & 0.909 & 0.720 & 0.566 & 0.263 & 0.990 & 0.400 & 0.825 & 0.641 \\
		  C+S Repair & 0.909 & 0.780 & 0.555 & 0.264 & 0.990 & 0.460 & 0.813 & 0.609 \\
				  \midrule
					  \multicolumn{9}{l}{\textbf{Baselines (Qwen3; no archive)}} \\
					  Qwen3 Sample & 0.387 & 0.242 & 0.896 & 0.876 & 0.300 & 0.200 & 0.939 & 0.952 \\
					  Qwen3 + Compile Repair & 0.371 & 0.237 & 0.912 & 0.911 & 0.900 & 0.260 & 0.740 & 0.385 \\
					  Qwen3 + C+S Repair & 0.688 & 0.548 & 0.802 & 0.656 & 0.940 & 0.390 & 0.827 & 0.655 \\
			  \midrule
			  \multicolumn{9}{l}{\textbf{Hybrid control (no archive)}} \\
						  Hybrid control & 0.973 & 0.828 & 0.505 & 0.241 & 1.000 & 0.530 & 0.790 & 0.588 \\
				  \midrule
			  \multicolumn{9}{l}{\textbf{FormalEvolve}} \\
						  FormalEvolve & 0.973 & 0.849 & 0.443 & 0.229 & 1.000 & \textbf{0.580} & 0.759 & 0.531 \\
					  FormalEvolve (K=1) & \textbf{0.984} & 0.866 & \textbf{0.362} & \textbf{0.198} & 1.000 & 0.550 & \textbf{0.726} & \textbf{0.442} \\
					  FormalEvolve w/o EvolAST & 0.973 & \textbf{0.871} & 0.454 & 0.236 & 1.000 & 0.500 & 0.776 & 0.577 \\
					  FormalEvolve w/o patch repair & 0.903 & 0.780 & 0.449 & 0.211 & 1.000 & 0.470 & 0.772 & 0.494 \\
			  \bottomrule
			  \end{tabular}
}%
\end{table*}

\paragraph{Ablations and component effects.}
The strongest Kimina-only baseline is C+S Repair.
Adding the stronger Qwen3 repair model without an archive already improves SH@100 from 0.460 to 0.530 on CombiBench and from 0.780 to 0.828 on ProofNet.
FormalEvolve then adds archive-based search on top of this repair strength, so the hybrid row can be read as a matched no-archive ablation.
Removing patch repair drops SH@100 from 0.580 to 0.470 on CombiBench and from 0.849 to 0.780 on ProofNet, identifying bounded repair as the largest direct driver of SH@100.
The archive then makes repaired candidates reusable as parents and prompt context, allowing local repair gains to propagate through later search steps while reducing repeated edits of the same templates.
The island and EvolAST ablations expose a dataset-dependent exploration--exploitation trade-off under the fixed budget.
\textbf{ProofNet (high-prior exploitation).} ProofNet consists of standard undergraduate mathematics, and the public autoformalization SFT mixture used to initialize Kimina-Autoformalizer includes ProofNet, together with PutnamBench, miniF2F, and Compfiles \citep{wang2025kimina}.
The Kimina-based generation component therefore starts with a particularly strong prior on this benchmark.
When the initial structures are already strong, centralizing the archive with $K{=}1$ or relying on LLM patch/repair without EvolAST can exploit a productive trajectory efficiently; additional island separation or structural perturbation may instead spend the fixed budget away from that trajectory.
This is consistent with the higher ProofNet SH@100 of $K{=}1$ (0.866) and no-EvolAST (0.871), compared with 0.849 for the default.
\textbf{CombiBench (domain-shift exploration).} CombiBench presents a substantially stronger domain/style shift, where model priors more often fail and standard patching can collapse into repeated, unhelpful edits.
Here, isolated search trajectories help prevent premature convergence, while EvolAST forces structural variation when LLM-driven repair stalls.
Consequently, $K{=}2$ with EvolAST reaches SH@100 of 0.580, compared with 0.550 for $K{=}1$ and 0.500 without EvolAST.
We therefore use $K{=}2$ with EvolAST as a robust default across different domain difficulties rather than as a configuration that is optimal for every dataset and metric; the main controlled comparison remains archive-based search versus matched no-archive controls.

\begin{figure*}[t]
  \centering
  \includegraphics[trim=6pt 6pt 6pt 6pt,clip,width=0.88\textwidth]{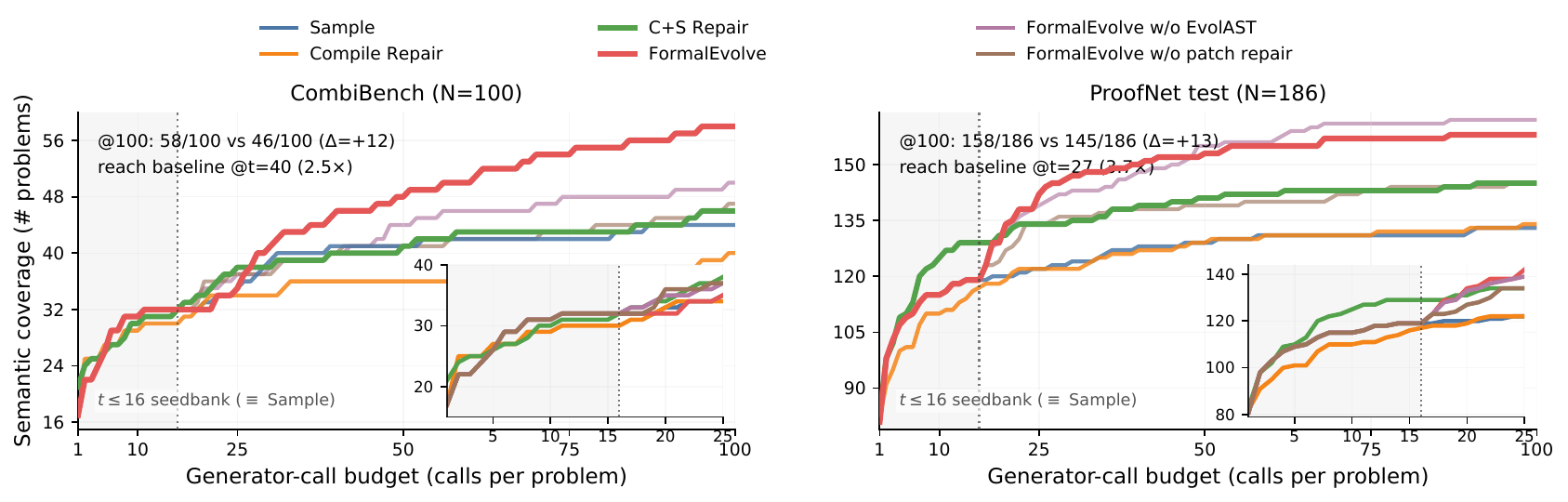}
  \caption{Semantic coverage vs debited generator calls $t$ on both benchmarks. The dotted line marks the nominal seedbank boundary ($t{=}16$).}
  \label{fig:main-curves}
\end{figure*}

\begin{figure*}[t]
  \centering
  \includegraphics[width=0.95\textwidth,trim=0pt 6pt 6pt 6pt,clip]{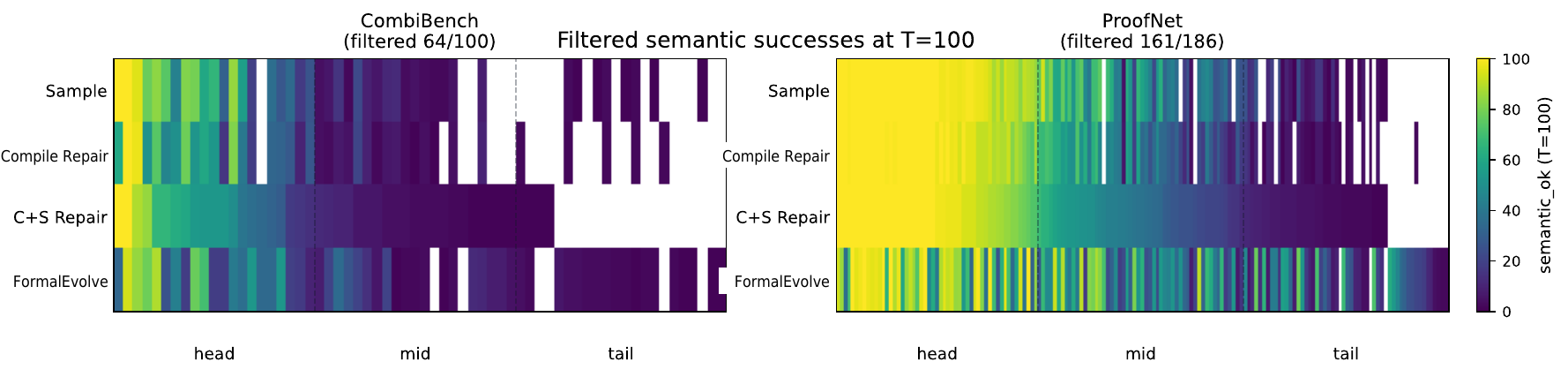}
  \caption{Per-problem semantic-success counts at $T=100$ (\texttt{semantic\_ok}; zeros shown as white). Columns are ordered by the C+S Repair baseline, and only problems with a positive count for at least one method are shown. The head/mid/tail regions are visualization bands induced by this baseline ordering, not predefined difficulty splits.}
  \label{fig:uniformity-filtered}
\end{figure*}

\paragraph{Cross-problem concentration.}
Coverage alone can hide a failure mode in which a method spends most of its successful calls on the same easy problems.
We therefore report Gini and top-10\% share over per-problem \texttt{semantic\_ok} counts.
The default FormalEvolve configuration lowers both concentration metrics relative to the hybrid control: on CombiBench, Gini decreases from 0.790 to 0.759 and top-10\% share from 0.588 to 0.531; on ProofNet, the corresponding changes are 0.505 to 0.443 and 0.241 to 0.229.
Bootstrap intervals over problems show that the Gini reductions are stable for the matched hybrid comparison (+0.031 [0.010, 0.055] on CombiBench; +0.062 [0.035, 0.090] on ProofNet), while top-10\% reductions have the same sign but wider intervals (Appendix~\ref{sec:appendix-light-exp}).
Under the C+S Repair ordering used in \Cref{fig:uniformity-filtered}, FormalEvolve's additional successes concentrate in the baseline-ranked mid and tail.
In the tail segment, FormalEvolve has higher semantic-success counts on 18 versus 2 lower-count problems on CombiBench, and on 49 versus 5 on ProofNet.
The head can still favor C+S Repair, especially on ProofNet, so the pattern reflects complementary problem-level coverage across the baseline-induced bands.
\Cref{fig:uniformity-filtered,fig:uniformity-gini} show the same pattern at the problem level and across budgets.

\subsection{Downstream Proof Utility}

\paragraph{Prover protocol.}
For each non-empty deduplicated semantic repertoire, Goedel-Prover-V2-32B receives $B=64$ attempts under a fixed round-robin schedule; candidates are revisited when the repertoire contains fewer than 64 candidates. Empty repertoires count as failures.
The single-reference GT control uses the dataset reference statement and therefore retries that statement for all 64 attempts.
Table~\ref{tab:proof-utility-64} reports solved-problem counts for pass@64, complete@64, and theorem-complete@64.
The prover sees only candidates with \texttt{compile\_ok}$=1$ and \texttt{semantic\_ok}$=1$, deduplicated under canonicalization.
The GT reference row runs the same prover on the dataset-provided formal statement and bypasses both generator and semantic judge.

\begin{table*}[t]
  \centering
  \caption{Downstream proof utility at $B=64$ prover attempts per problem. Theorem-complete counts complete proofs of explicit \texttt{theorem}/\texttt{lemma} declarations.}
  \label{tab:proof-utility-64}
  \scriptsize
  \setlength{\tabcolsep}{5pt}
  \resizebox{\textwidth}{!}{%
  \begin{tabular}{lccc|ccc}
  \toprule
   & \multicolumn{3}{c}{CombiBench ($N=100$)} & \multicolumn{3}{c}{ProofNet (test, $N=186$)} \\
  Method & pass@64 & complete@64 & thm-complete@64 & pass@64 & complete@64 & thm-complete@64 \\
  \midrule
  FormalEvolve & \PUCBPassOurs & \PUCBCompleteOurs & \PUCBTheoremCompleteOurs & \PUPNPassOurs & \PUPNCompleteOurs & \PUPNTheoremCompleteOurs \\
  Hybrid control & \PUCBPassHybrid & \PUCBCompleteHybrid & \PUCBTheoremCompleteHybrid & \PUPNPassHybrid & \PUPNCompleteHybrid & \PUPNTheoremCompleteHybrid \\
  Sample & \PUCBPassSample & \PUCBCompleteSample & \PUCBTheoremCompleteSample & \PUPNPassSample & \PUPNCompleteSample & \PUPNTheoremCompleteSample \\
  C+S Repair & \PUCBPassStrongSemantic & \PUCBCompleteStrongSemantic & \PUCBTheoremCompleteStrongSemantic & \PUPNPassStrongSemantic & \PUPNCompleteStrongSemantic & \PUPNTheoremCompleteStrongSemantic \\
  \midrule
  GT reference & \PUCBPassGT & \PUCBCompleteGT & \PUCBTheoremCompleteGT & \PUPNPassGT & \PUPNCompleteGT & \PUPNTheoremCompleteGT \\
  \bottomrule
  \end{tabular}
  }%
\end{table*}

\paragraph{Theorem-complete gains and boundaries.}
On CombiBench, FormalEvolve raises theorem-complete@64 from 9/100 for the matched Hybrid control to 13/100, and from 8/100 for both Sample and Kimina C+S Repair.
On ProofNet, FormalEvolve also improves over the matched Hybrid control on theorem-complete@64 (45/186 vs.\ 40/186), while remaining comparable to Kimina C+S Repair (45/186 vs.\ 46/186).
These results are consistent with the hedging motivation: diverse repertoires can increase the chance of reaching a prover-friendly accepted statement under a fixed prover budget.
Appendix~\ref{sec:appendix-proof-utility-decomposition} and Appendix~\ref{sec:appendix-prover-audit} provide additional curves and audit details.

\subsection{Test-Time Cost and Efficiency}
\label{sec:cost-efficiency}

\begin{figure*}[t]
  \centering
  \includegraphics[width=0.90\textwidth]{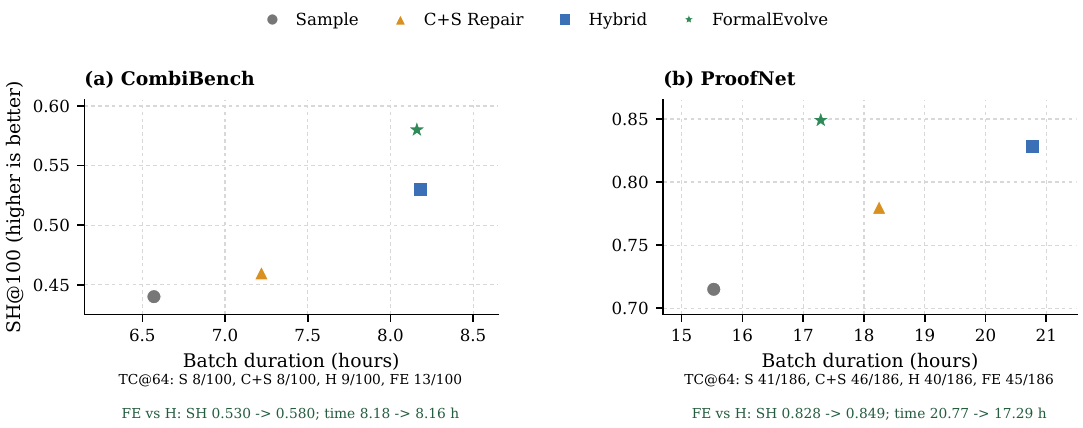}
  \caption{SH@100 versus batch duration under the same $T{=}100$ generator-call budget. The footer beneath each panel reports theorem-complete@64 and the FormalEvolve--Hybrid comparison.}
  \label{fig:cost-performance}
\end{figure*}

Under the same $T{=}100$ generator-call budget, FormalEvolve improves both SH@100 and theorem-complete@64 over the matched Hybrid control, with end-to-end durations of 8.16 vs.\ 8.18 hours on CombiBench and 17.29 vs.\ 20.77 hours on ProofNet.
\Cref{fig:cost-performance} presents this final-result comparison.
Appendix~\ref{sec:appendix-budget-audit} reports the complete accounting of generation, compilation-repair, and semantic-repair calls, semantic-judge calls, Lean evaluations, candidate records, and batch duration.

\subsection{Stronger-Base and Manual Audits}

\noindent\textbf{Stronger-base statement-generation.} As a supplementary CombiBench statement-generation check, we use Goedel-Formalizer-V2-8B seeding and GPT-5.4 repair with the same judge and $T{=}100$ accounting.
FormalEvolve preserves the pattern over the matched Hybrid control: SH@100 rises from 0.86 to 0.88, deduplicated SemOK$_\text{total}$ rises from 1886 to 3238 (1.7$\times$), and concentration drops in both Gini (0.4832 to 0.4383) and top-10\% share (0.3128 to 0.2202) (Appendix~\ref{sec:appendix-stronger-base-eval}).
\par\noindent\textbf{Manual faithfulness calibration.} Manual audits calibrate selected statement-stage and theorem-complete slices.
On a matched 50-problem ProofNet statement-stage audit, FormalEvolve, Hybrid control, and Sample have similar Faithful/Partial/Severe totals: 42/5/3, 43/4/3, and 43/3/4.
These similar profiles calibrate FormalEvolve's judge-positive outputs against the no-archive controls in this slice.
At the theorem-complete end, a 161-artifact audit over the original three-method pool shows that FormalEvolve yields the largest absolute count of faithful artifacts (41), while severe counts are numerically similar across methods (11, 10, and 12).
Proof success establishes utility for the encoded formal statement under the fixed prover protocol.
Manual auditing separately assesses faithfulness to the source statement (Appendix~\ref{sec:appendix-manual-faithfulness-audit}).

\section{Conclusion}
\label{sec:conclusion}

FormalEvolve is a compile-gated test-time search method for Lean~4 autoformalization under a fixed generator-call budget.
The central lesson is that autoformalization should be evaluated at the level of budgeted, semantically filtered repertoires: faithful variants of the same informal claim can expose different prover-facing interfaces.
FormalEvolve operationalizes this view with a compilation-feasible archive that stores reusable candidates, expands them through repair and evolution, and filters the resulting repertoire for downstream proving.
Across two benchmarks, this archive-based organization improves semantic coverage, reduces cross-problem concentration, and yields theorem-complete gains over matched no-archive multi-candidate controls under our fixed prover protocol.
The same archive-and-filter pattern may also extend to Coq, Isabelle, and related proof assistants, where multiple faithful formalizations can likewise differ in downstream prover utility.

\section*{Limitations}
\label{sec:limitations}

Semantic consistency remains difficult to certify automatically, which is a central challenge for autoformalization.
Our semantic metrics rely on an LLM-as-judge implementation of $\SemOK$, so results that use semantic labels--including semantic coverage, semantic-success counts, and concentration statistics--can depend on the judge model, prompt, and serving details.
Compilation certifies Lean elaboration and type correctness, while semantic equivalence to the informal problem remains outside the compiler's guarantee.
We therefore report manual faithfulness audits over outputs produced by our pipeline.
On the matched ProofNet statement-stage slice, FormalEvolve, Hybrid control, and Sample have similar Faithful/Partial/Severe totals (42/5/3, 43/4/3, and 43/3/4), suggesting that archive search does not visibly amplify judge-positive semantic shift relative to no-archive controls.

Our pass@64, complete@64, and theorem-complete@64 results use one fixed prover configuration with $B=64$ attempts.

\section*{Acknowledgments}
This work was primarily conducted at the State Key Laboratory of General Artificial Intelligence, BIGAI.

\bibliography{references}

\clearpage
\appendix

\makeatletter
\@addtoreset{algorithm}{section}
\makeatother
\setcounter{algorithm}{0}
\renewcommand{\thealgorithm}{\thesection.\arabic{algorithm}}
\renewcommand{\theHalgorithm}{\thesection.\arabic{algorithm}}

\section{Additional Results}
\label{sec:appendix-additional-results}

We give the metric definitions used throughout the experiments, followed by statement-level decompositions, proof-utility decomposition, and budget-sweep coverage gaps under the same generator-call accounting.

\subsection{Metric definitions}
\label{sec:appendix-metric-definitions}
We provide definitions of all reporting metrics under generator-call accounting.
For a given problem, let $E_t$ denote the set of candidates evaluated up to debited call $t$, and let $E_t^{\mathrm{feas}}=\{c\in E_t : \CompOK(c)=1\}$.
Let $G_t=\{c\in E_t : \CompOK(c)=1 \wedge \SemOK(c)=1\}$, and for dataset-level concentration let $s_j(t)$ be the number of semantic-success candidates for problem $j$ at budget $t$.
For proof utility, let $\tilde{G}_t=\mathrm{Dedup}(G_t)$ be the deduplicated semantic repertoire and let $\mathcal{A}_t^{(64)}=\mathrm{RR64}(\tilde{G}_t)$ be its ordered sequence of $B=64$ prover attempts.
For each problem with a non-empty repertoire, RR64 allocates 64 attempts in round-robin order over the available candidates; if the repertoire contains fewer than 64 candidates, candidates are revisited until the attempt budget is exhausted.
The GT reference contains one statement and is therefore retried for all 64 attempts.
Let $P_{\mathrm{pass}}(a)\in\{0,1\}$ indicate whether prover attempt $a$ returns a Lean~4-accepted proof script (warnings allowed; \texttt{sorry} permitted), $P_{\mathrm{complete}}(a)\in\{0,1\}$ indicate a complete proof without \texttt{sorry}, and $P_{\mathrm{theorem}}(a)\in\{0,1\}$ indicate a complete proof of an explicit \texttt{theorem}/\texttt{lemma}.

\begin{table}[H]
  \centering
  \caption{Metric definitions under generator-call accounting.}
  \label{tab:metrics-at-a-glance}
  \scriptsize
  \begingroup
  \renewcommand{\arraystretch}{1.02}
  \setlength{\tabcolsep}{2pt}
  \hyphenpenalty=2000
  \exhyphenpenalty=2000
  \begin{tabular}{@{}>{\raggedright\arraybackslash}p{0.23\columnwidth}@{\hspace{4pt}}>{\raggedright\arraybackslash}p{0.72\columnwidth}@{}}
	    \toprule
	    Metric & Definition and interpretation \\
	    \midrule
	    $\mathrm{FY}(t)$ & $\frac{1}{|E_t|}\sum_{c\in E_t} \CompOK(c)$; fraction of debited calls yielding compilable candidates. \\
	    $\mathrm{CH}(t)$ & $I[\exists c \in E_t : \CompOK(c)=1]$; at least one compilable candidate. \\
	    $\mathrm{SH}(t)$ & $I[\exists c \in E_t : \CompOK(c)=1 \wedge \SemOK(c)=1]$; at least one compilable and judge-consistent candidate. \\
	    $\mathrm{SD}(t)$ & $\frac{1}{|E_t^{\mathrm{feas}}|}\sum_{c\in E_t^{\mathrm{feas}}} \SemOK(c)$, or 0 if $E_t^{\mathrm{feas}}$ is empty; semantic density among compilable candidates. \\
	    $\mathrm{SY}(t)$ & $\frac{1}{|E_t|}\sum_{c\in E_t} \SemOK(c)$; semantic yield per call, treating non-compilable candidates as $\SemOK=0$. \\
	    $\mathrm{Div}(t)$ & $|\mathrm{Dedup}(G_t)|$; deduplicated semantic successes. \\
	    $\mathrm{Cov}(t)$ & $\frac{1}{N}\sum_{j=1}^N I[s_j(t)\ge 1]$; dataset-level semantic coverage. \\
	    $\mathrm{Gini}(t)$ & $\frac{\sum_{i,j}|s_i(t)-s_j(t)|}{2N\sum_j s_j(t)+\varepsilon}$; concentration of semantic successes ($\downarrow$). \\
	    Top-10\% & $\frac{\sum_{j\in\mathrm{Top}_{10\%}}s_j(t)}{\sum_j s_j(t)}$; success mass assigned to the easiest 10\% problems ($\downarrow$). \\
    PU-pass@64 & At least one RR64 prover attempt returns a Lean~4-accepted proof script. \\
    PU-complete@64 & At least one RR64 prover attempt returns a proof without \texttt{sorry}. \\
    PU-theorem@64 & At least one RR64 prover attempt returns a complete proof of an explicit \texttt{theorem}/\texttt{lemma}. \\
	    \midrule
	    \multicolumn{2}{>{\raggedright\arraybackslash}p{0.96\columnwidth}}{Each debited generator call yields one evaluated representative candidate; repair and semantic-repair calls count as additional calls. Dedup uses conservative whitespace/name canonicalization.} \\
	    \bottomrule
	  \end{tabular}
	  \endgroup
\end{table}

\subsection{Proof utility decomposition}
\label{sec:appendix-proof-utility-decomposition}
\begin{figure*}[!b]
  \centering
  \includegraphics[trim=6pt 6pt 6pt 6pt,clip,width=0.96\textwidth]{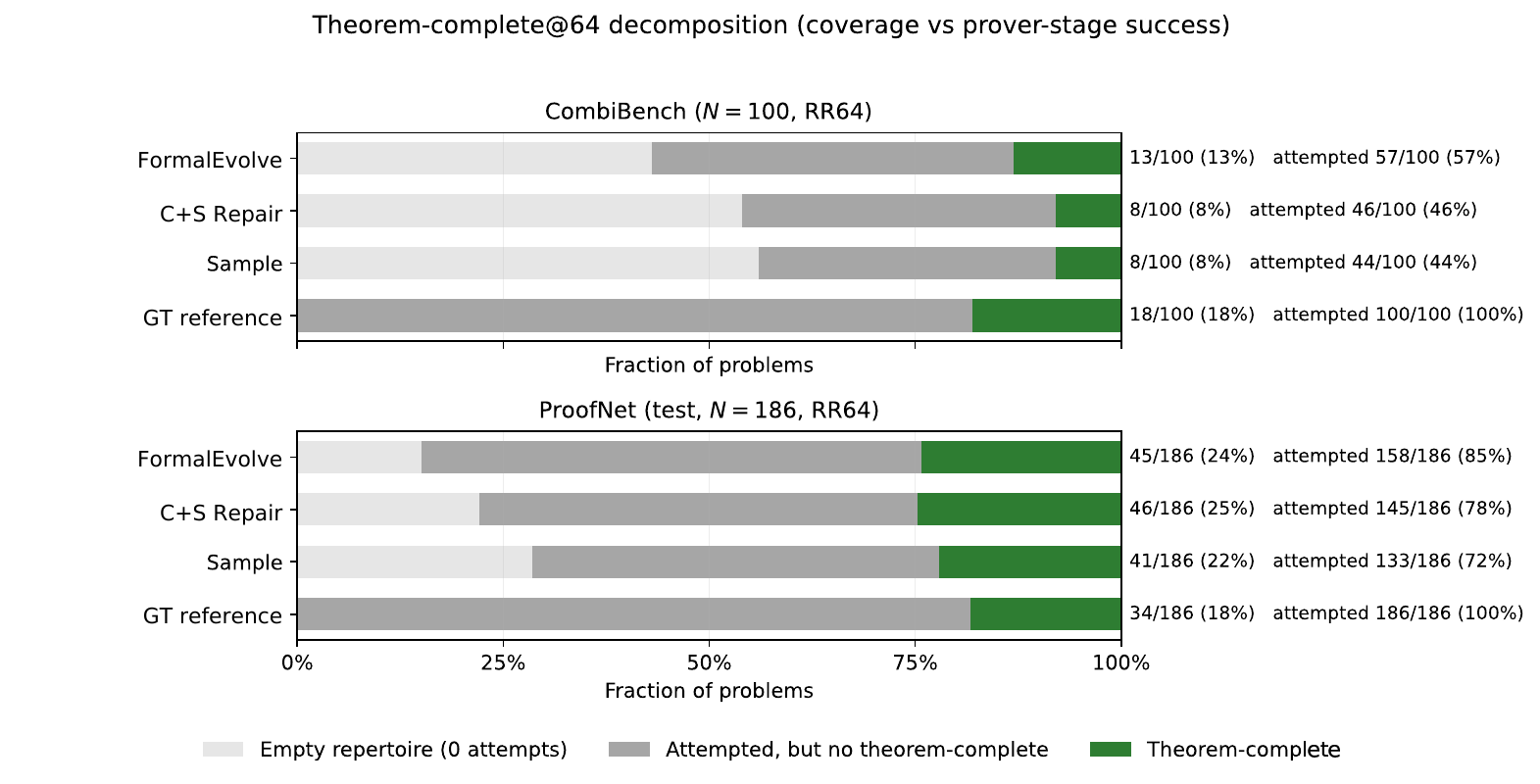}
  \caption{Decomposing theorem-complete@64 under a fixed prover and attempt budget ($B{=}64$). Each bar splits the benchmark denominator into problems with an empty statement repertoire for proving (zero prover attempts), problems with a non-empty repertoire but no theorem-complete success, and problems with at least one theorem-complete success.}
  \label{fig:proof-utility-decomposition}
\end{figure*}

Figure~\ref{fig:proof-utility-decomposition} decomposes theorem-complete@64 into (i) whether a method produces a non-empty statement repertoire to attempt under the fixed prover protocol, and (ii) whether at least one attempted statement is solved at theorem-complete level under the same attempt budget.

\subsection{\texorpdfstring{Statement-level decompositions at $T=100$}{Statement-level decompositions at T=100}}
Tables~\ref{tab:main-100} and \ref{tab:proofnet-100} decompose the fixed-budget statement-level results.
FY is compilation yield, SD is semantic density conditional on compilation, and SY is semantic yield per generator call (treating compilation failures as semantic failures).
Reporting FY/SH/SD/SY together separates gains from (i) producing more compilable candidates, (ii) improving semantic consistency among compilable candidates, and (iii) reallocating budget between proposal and repair.

\begin{table}[H]
  \centering
  \caption{CombiBench statement-level results at $T=100$.}
  \label{tab:main-100}
  \scriptsize
  \resizebox{\columnwidth}{!}{%
  \begin{tabular}{lcccc}
  \toprule
  Method & FY & SH & SD & SY \\
  \midrule
  Sample & 0.695 & 0.440 & 0.189 & 0.132 \\
  Compile Repair & 0.555 & 0.400 & 0.199 & 0.110 \\
  C+S Repair & 0.324 & 0.460 & 0.371 & 0.120 \\
  FormalEvolve & 0.407 & 0.580 & 0.326 & 0.133 \\
  FormalEvolve w/o EvolAST & 0.346 & 0.500 & 0.313 & 0.108 \\
  FormalEvolve w/o patch repair & 0.682 & 0.470 & 0.250 & 0.170 \\
  \bottomrule
  \end{tabular}
  }
\end{table}

\begin{table}[H]
  \centering
  \caption{ProofNet statement-level results at $T=100$.}
  \label{tab:proofnet-100}
  \scriptsize
  \resizebox{\columnwidth}{!}{%
  \begin{tabular}{lcccc}
  \toprule
  Method & FY & SH & SD & SY \\
  \midrule
  Sample & 0.702 & 0.715 & 0.599 & 0.421 \\
  Compile Repair & 0.625 & 0.720 & 0.622 & 0.389 \\
  C+S Repair & 0.509 & 0.780 & 0.760 & 0.387 \\
  FormalEvolve & 0.562 & 0.849 & 0.775 & 0.435 \\
  FormalEvolve w/o EvolAST & 0.539 & 0.871 & 0.788 & 0.425 \\
  FormalEvolve w/o patch repair & 0.694 & 0.780 & 0.684 & 0.475 \\
  \bottomrule
  \end{tabular}
  }
\end{table}

\subsection{Coverage gaps across the budget}
\begin{figure*}[!b]
  \centering
  \includegraphics[trim=6pt 6pt 6pt 6pt,clip,width=\textwidth]{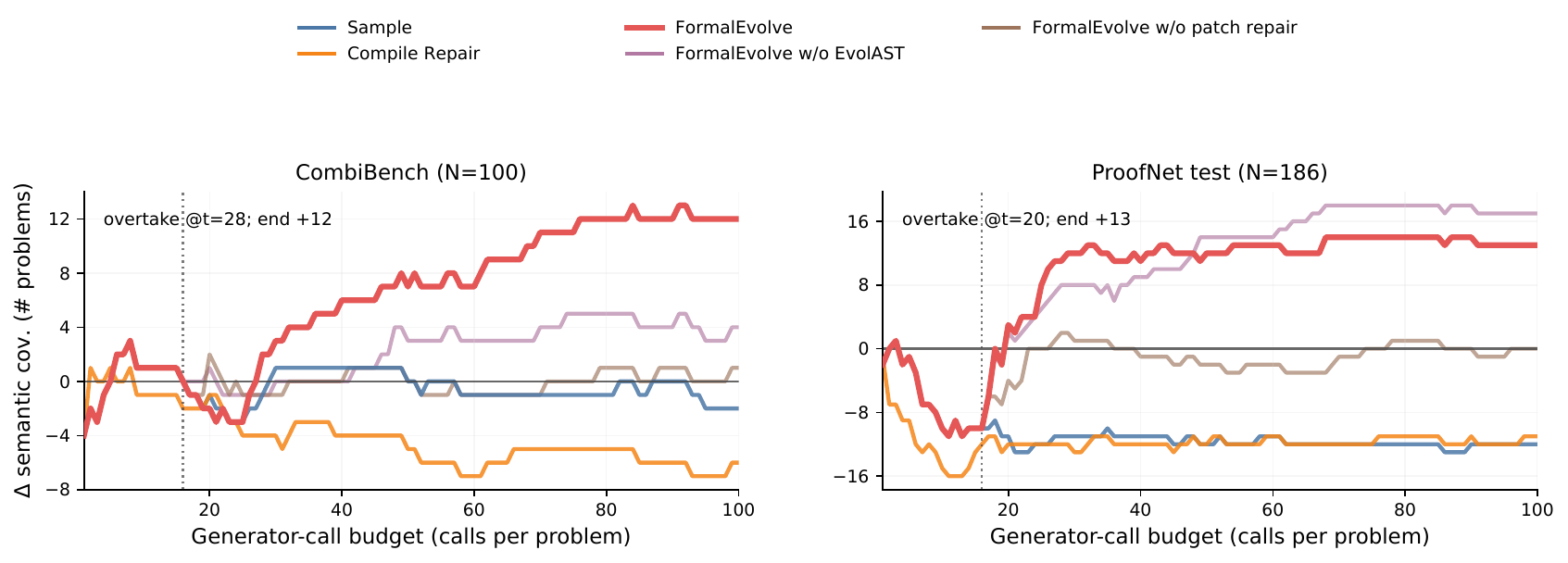}
\caption{Coverage gap vs debited calls, plotted as $\Delta$ semantic coverage relative to the strongest Kimina sampling baseline (Compile+Semantic Repair). The zero line corresponds to the baseline; positive values indicate a method covers more problems at the same generator-call budget. The dotted vertical line marks the nominal seedbank boundary ($t{=}16$) used by evolution-based methods (pre-sampled seeds, debited calls); a small subset of problems can require additional debited seeds before evolution starts.}
  \label{fig:delta-coverage}
\end{figure*}

Figure~\ref{fig:delta-coverage} reports the budget-sweep coverage gap relative to the strongest Kimina sampling baseline.
The hybrid Kimina+Qwen3 baseline in Table~\ref{tab:full-hit100} separates search effects from patch-model strength under the same generator-call accounting and repair protocol as the sampling baselines.

\clearpage

\section{Analysis}
\label{sec:analysis}

We provide additional measurements for the fixed-budget experiments: paired uncertainty, repair-call composition, prover-stage non-dominance, cross-problem concentration, stronger-base behavior, and component effects.


\subsection{Paired uncertainty, call composition, and prover non-dominance}
\label{sec:appendix-light-exp}
The measurements below are computed from the same fixed-budget artifacts used for the main figures and tables.
They cover paired uncertainty (Table~\ref{tab:appendix-sec5-ci}), concentration uncertainty for the matched hybrid comparison (Table~\ref{tab:appendix-sec5-concentration-ci}), per-problem repair-call composition (Table~\ref{tab:appendix-sec5-call-comp}), and prover-stage non-dominance (Table~\ref{tab:appendix-sec5-nondom}).

\begin{table}[H]
  \centering
  \caption{Paired uncertainty over problems for the matched Hybrid no-archive comparison. Hybrid denotes the matched Kimina-generation/Qwen3-repair no-archive control used in Table~\ref{tab:full-hit100} and Table~\ref{tab:proof-utility-64}.}
  \label{tab:appendix-sec5-ci}
  \scriptsize
  \setlength{\tabcolsep}{3pt}
  \resizebox{\columnwidth}{!}{%
  \begin{tabular}{lcccc}
    \toprule
    Dataset & Metric & mean(FE) & mean(base) & $\Delta$ [95\% CI] \\
    \midrule
    CombiBench & SH@100 vs Hybrid & 0.580 & 0.530 & +0.050 [0.003, 0.130] \\
    ProofNet & SH@100 vs Hybrid & 0.849 & 0.828 & +0.022 [0.005, 0.048] \\
    \bottomrule
  \end{tabular}
  }
\end{table}
We use paired bootstrap 95\% confidence intervals for the mean difference (FormalEvolve $-$ baseline).

\begin{table}[H]
  \centering
  \caption{Bootstrap uncertainty for concentration reductions relative to the matched hybrid no-archive control. Positive $\Delta$ means the hybrid control is more concentrated than FormalEvolve, so lower concentration is better for FormalEvolve.}
  \label{tab:appendix-sec5-concentration-ci}
  \scriptsize
  \setlength{\tabcolsep}{3pt}
  \resizebox{\columnwidth}{!}{%
  \begin{tabular}{lccccc}
    \toprule
    Dataset & Metric & FE & Hybrid & $\Delta$ (Hybrid $-$ FE) & 95\% CI \\
    \midrule
    CombiBench & Gini & 0.759 & 0.790 & +0.031 & [0.010, 0.055] \\
    CombiBench & Top-10\% share & 0.531 & 0.588 & +0.057 & [-0.009, 0.128] \\
    ProofNet & Gini & 0.443 & 0.505 & +0.062 & [0.035, 0.090] \\
    ProofNet & Top-10\% share & 0.229 & 0.241 & +0.011 & [-0.005, 0.032] \\
    \bottomrule
  \end{tabular}
  }
\end{table}
We compute the intervals by resampling benchmark problems with replacement and recomputing each concentration metric on the resampled per-problem semantic-success counts.

\begin{table}[H]
  \centering
  \caption{Debited repair-call composition under $T=100$.}
  \label{tab:appendix-sec5-call-comp}
  \scriptsize
  \setlength{\tabcolsep}{3pt}
  \resizebox{\columnwidth}{!}{%
  \begin{tabular}{l l r r}
    \toprule
    Dataset & Method & CRep (median [q25,q75]) & SRep (median [q25,q75]) \\
    \midrule
    CombiBench & Sample & 0.0 [0.0, 0.0] & 0.0 [0.0, 0.0] \\
    CombiBench & Compile Repair & 29.0 [4.0, 52.0] & 0.0 [0.0, 0.0] \\
    CombiBench & C+S Repair & 14.0 [4.0, 34.5] & 38.0 [26.0, 54.0] \\
    CombiBench & Hybrid control & 15.5 [2.8, 33.0] & 43.5 [25.8, 57.0] \\
    CombiBench & FormalEvolve & 30.5 [20.8, 39.0] & 20.5 [14.0, 26.0] \\
    ProofNet & Sample & 0.0 [0.0, 0.0] & 0.0 [0.0, 0.0] \\
    ProofNet & Compile Repair & 16.0 [0.0, 53.0] & 0.0 [0.0, 0.0] \\
    ProofNet & C+S Repair & 9.0 [0.0, 39.8] & 11.5 [0.0, 37.8] \\
    ProofNet & Hybrid control & 6.0 [0.0, 29.0] & 10.0 [0.0, 32.8] \\
    ProofNet & FormalEvolve & 24.0 [8.0, 40.8] & 4.0 [0.0, 16.0] \\
    \bottomrule
  \end{tabular}
  }
\end{table}

\begin{table}[H]
  \centering
  \caption{Prover-stage non-dominance at theorem-complete@64 for FormalEvolve versus Compile+Semantic Repair (C+S Repair).}
  \label{tab:appendix-sec5-nondom}
  \scriptsize
  \setlength{\tabcolsep}{3pt}
  \resizebox{\columnwidth}{!}{%
  \begin{tabular}{l r r r r r}
    \toprule
    Dataset & n & both fail & FormalEvolve only & C+S Repair only & both succeed \\
    \midrule
    CombiBench & 100 & 85 & 7 & 2 & 6 \\
    ProofNet & 186 & 129 & 11 & 12 & 34 \\
    \bottomrule
  \end{tabular}
  }
\end{table}

\FloatBarrier

\subsection{Cross-problem uniformity: coverage and concentration}
We assess whether semantic successes concentrate on a small subset of easy problems under a fixed call budget.
At $T=100$, we summarize the per-problem \texttt{semantic\_ok} count distribution using coverage (SH@100) and concentration (Gini and top-10\% share; Table~\ref{tab:full-hit100}).
Figure~\ref{fig:uniformity-filtered} provides a compact per-problem view. For readability, we filter to problems where at least one method attains a positive \texttt{semantic\_ok} count (otherwise the column is identically zero).
We order columns by the strongest Kimina sampling baseline (Compile+Semantic Repair) using its per-problem \texttt{semantic\_ok} counts at $T=100$, defining an x-axis independent of FormalEvolve; Figure~\ref{fig:delta-strips} uses the same ordering for the delta visualization.
This subsection uses Compile+Semantic Repair only as a fixed ordering and visualization baseline; the matched Hybrid comparisons are reported separately in Table~\ref{tab:appendix-sec5-ci} and in the main results.

\paragraph{Aggregate pattern under baseline ordering.}
Under this baseline ordering, gains concentrate in the baseline-ranked tail: on the filtered set, $\Delta>0$ dominates in the tail segment, while the head segment can favor the baseline (especially on ProofNet).
We define head/mid/tail by splitting the baseline-ordered problems into three contiguous segments as evenly as possible within each benchmark.
The table below counts how often $\Delta=\#\texttt{semantic\_ok}(\textsc{FormalEvolve})-\#\texttt{semantic\_ok}(\textsc{Compile+Semantic Repair})$ is positive/negative/zero within each segment (reported as $+/-/0$).
\begin{center}
\small
\setlength{\tabcolsep}{6pt}
\resizebox{\columnwidth}{!}{%
\begin{tabular}{lccc}
\toprule
Dataset & head ($+/-/0$) & mid ($+/-/0$) & tail ($+/-/0$) \\
\midrule
CombiBench (filtered 64/100) & 8/12/1 & 12/7/2 & 18/2/2 \\
ProofNet (filtered 161/186) & 6/41/6 & 35/19/0 & 49/5/0 \\
\bottomrule
\end{tabular}
}
\end{center}

\subsubsection{Representative early semantic hits and baseline-miss instances}
\label{sec:appendix-early-wins}
There are 16 such instances on each benchmark, where Compile+Semantic Repair has zero \texttt{semantic\_ok} at $T{=}100$ while FormalEvolve attains a positive \texttt{semantic\_ok} count.
Conversely, there are 4 such instances where the baseline hits but FormalEvolve misses on CombiBench, and 3 on ProofNet.
To ground the aggregate patterns above and illustrate non-dominance, Table~\ref{tab:early-wins} lists representative instances in both directions: where FormalEvolve reaches a semantically consistent compilable statement substantially earlier than the baseline (or the baseline has no hit within $T=100$ calls), and where the baseline hits but FormalEvolve misses within the same call budget.
These examples are computed from the debited-call logs under the same budget.

\begin{table*}[t]
  \centering
\caption{Representative early semantic hits at a generator-call budget of $T=100$. We include both directions (FormalEvolve earlier / baseline no-hit, and baseline-hit / FormalEvolve no-hit). ``$>100$'' indicates no semantic hit within the call budget.}
  \label{tab:early-wins}
  \scriptsize
  \setlength{\tabcolsep}{3pt}
  \begin{tabular}{@{}>{\raggedright\arraybackslash}p{0.13\textwidth}>{\raggedright\arraybackslash}p{0.24\textwidth}>{\raggedright\arraybackslash}p{0.41\textwidth}>{\centering\arraybackslash}p{0.09\textwidth}>{\centering\arraybackslash}p{0.07\textwidth}@{}}
    \toprule
    Benchmark & Problem id & Short description & FormalEvolve first hit & C+S Repair first hit \\
    \midrule
    \multicolumn{5}{l}{\textbf{FormalEvolve hits earlier, or baseline has no hit within $T=100$}} \\
    CombiBench & \path{0055_egmo_2022_p5} & Domino tilings: parity of $f(n,2k)$ for all $k$ & 1 & $>100$ \\
    CombiBench & \path{0063_usamo_2000_p4} & 3 colored squares form an axis-aligned right triangle & 4 & $>100$ \\
    CombiBench & \path{0077_imo_2010_p5} & Box/coin operations reach $2010^{2010^{2010}}$? & 4 & $>100$ \\
    CombiBench & \path{0035_brualdi_ch9_8} & Count SDRs for six 2-sets arranged in a cycle & 21 & $>100$ \\
    CombiBench & \path{0011_brualdi_ch1_10} & No magic square of order $2$ & 22 & 93 \\
    CombiBench & \path{0047_brualdi_ch13_10} & Tournament: a vertex reaches all others within $2$ steps & 27 & 95 \\
    CombiBench & \path{0057_imosl_2015_c6} & Infinitely many integers without a unique odd-sum representation & 2 & 31 \\
    CombiBench & \path{0065_imo_2020_p3} & Split weighted colored pebbles into two equal piles & 27 & 52 \\
    \midrule
    ProofNet & \path{0046_exercise_3_4_5b} & Quotients of solvable groups are solvable & 19 & $>100$ \\
    ProofNet & \path{0124_exercise_26_12} & Perfect map: compact $Y$ implies compact $X$ & 21 & $>100$ \\
    ProofNet & \path{0163_exercise_3_13} & Cauchy product of absolutely convergent series & 25 & $>100$ \\
    ProofNet & \path{0119_exercise_23_3} & Connected union with shared intersection & 18 & 86 \\
    ProofNet & \path{0095_exercise_5_6_14} & Distinct roots of $x^m-x$ in characteristic $p$ & 24 & 91 \\
    ProofNet & \path{0045_exercise_3_4_4} & Finite abelian groups have subgroups of each divisor order & 12 & 62 \\
    ProofNet & \path{0005_exercise_6_4_12} & No simple group of order $224$ & 6 & 45 \\
    \midrule
    \multicolumn{5}{l}{\textbf{Baseline hits, but FormalEvolve has no hit within $T=100$}} \\
    CombiBench & \path{0037_brualdi_ch10_31} & Difference set $B=\{0,3,4,9,11\}$ in $\mathbb{Z}_{21}$ & $>100$ & 22 \\
    CombiBench & \path{0014_brualdi_ch2_36} & Counting combinations of multisets with bounded repetitions & $>100$ & 39 \\
    CombiBench & \path{0068_imo_2000_p4} & Distribute cards into 3 boxes so sums are equal & $>100$ & 50 \\
    CombiBench & \path{0021_brualdi_ch4_9} & Max inversions in a permutation of $\{1,\dots,n\}$ & $>100$ & 85 \\
    \midrule
    ProofNet & \path{0185_exercise_5_1} & Blaschke condition for zeros of bounded holomorphic functions & $>100$ & 3 \\
    ProofNet & \path{0009_exercise_11_2_13} & Divisibility in Gaussian integers implies divisibility in $\mathbb{Z}$ & $>100$ & 40 \\
    ProofNet & \path{0094_exercise_5_4_3} & A root of a polynomial with radicals is algebraic & $>100$ & 49 \\
    \bottomrule
  \end{tabular}
\end{table*}

\begin{figure*}[t]
  \centering
  \includegraphics[trim=6pt 6pt 6pt 6pt,clip,width=0.95\textwidth]{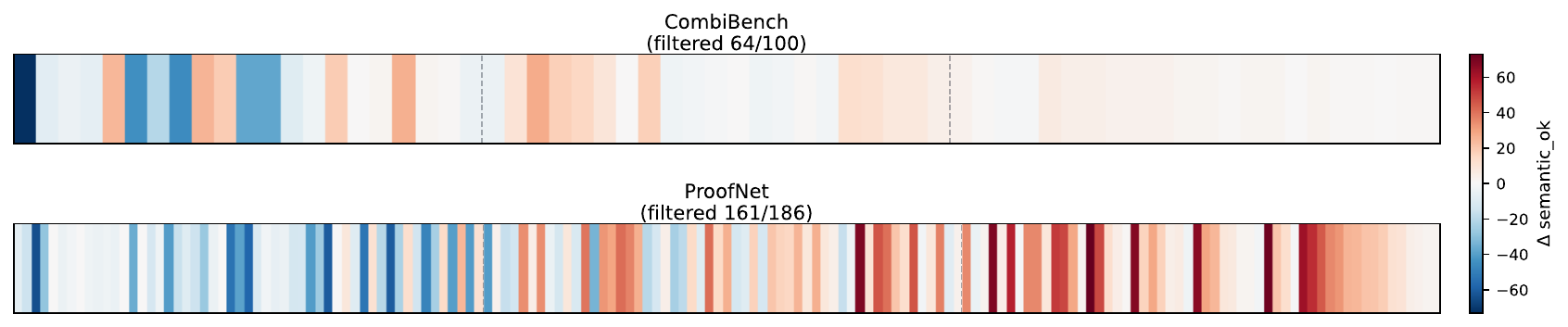}
\caption{Delta strips at $T=100$ under fixed-budget accounting on the same filtered set and in the same baseline ordering (sorted by Compile+Semantic Repair \texttt{semantic\_ok}) as Figure~\ref{fig:uniformity-filtered}, showing $\Delta=\#\texttt{semantic\_ok}(\textsc{FormalEvolve})-\#\texttt{semantic\_ok}(\textsc{Compile+Semantic Repair})$ per problem.}
  \label{fig:delta-strips}
\end{figure*}

\begin{figure*}[t]
  \centering
  \includegraphics[trim=6pt 6pt 6pt 6pt,clip,width=\textwidth]{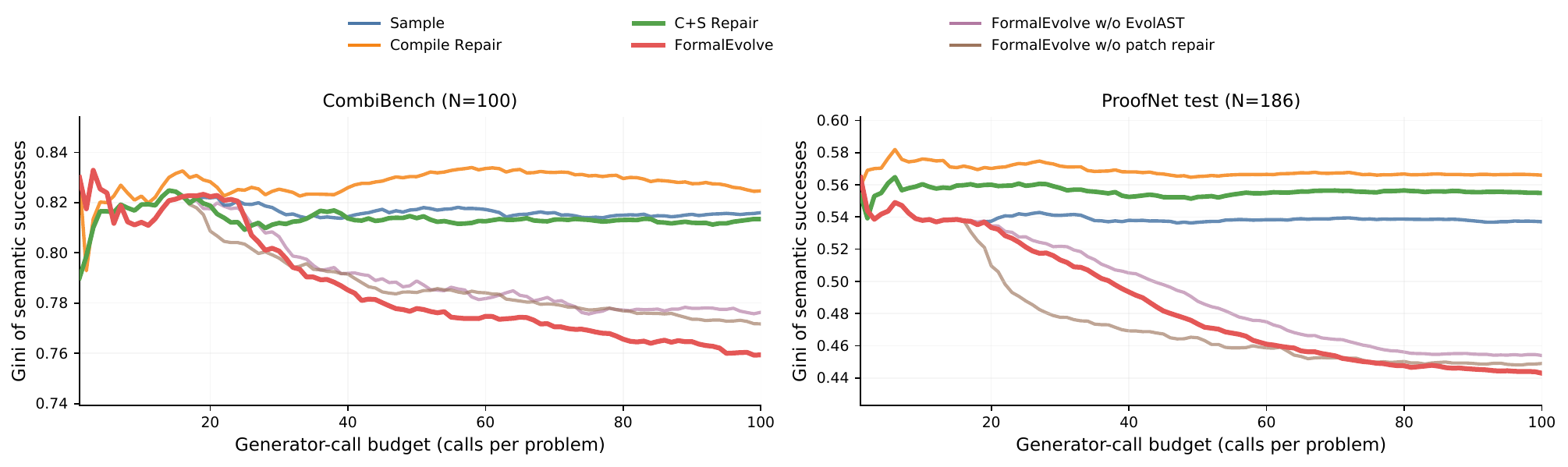}
  \caption{Cross-problem concentration of semantic successes vs generator-call budget, summarized by the Gini coefficient of per-problem semantic-success counts under fixed-budget accounting. Lower is more uniform (less concentrated on a small subset of problems). Table~\ref{tab:full-hit100} reports the corresponding summary at $T=100$.}
  \label{fig:uniformity-gini}
\end{figure*}

Across both benchmarks, FormalEvolve improves coverage and reduces concentration (see Table~\ref{tab:full-hit100} and Figures~\ref{fig:uniformity-filtered}--\ref{fig:uniformity-gini}).
\FloatBarrier

\subsection{Stronger-base statement-generation evaluation on CombiBench}
\label{sec:appendix-stronger-base-eval}

\begin{table}[H]
  \centering
  \caption{Stronger-base statement-generation evaluation on CombiBench ($N{=}100$, $T{=}100$). SemOK$_\text{total}$ is the deduplicated aggregate count of judge-consistent compiled candidates across problems.}
  \label{tab:stronger-base-eval}
  \scriptsize
  \setlength{\tabcolsep}{3pt}
  \resizebox{\columnwidth}{!}{%
  \begin{tabular}{lccccc}
    \toprule
    Method & CH@100 & SH@100 & SemOK$_\text{total}$ & Gini$\downarrow$ & Top-10\%$\downarrow$ \\
    \midrule
    Sample & 0.95 & 0.82 & 1478 & 0.5852 & 0.3410 \\
    C+S Repair & 0.93 & 0.85 & 1443 & 0.5693 & 0.3389 \\
    Hybrid control & \textbf{0.98} & 0.86 & 1886 & 0.4832 & 0.3128 \\
    \textbf{FormalEvolve} & 0.95 & \textbf{0.88} & \textbf{3238} & \textbf{0.4383} & \textbf{0.2202} \\
    \bottomrule
  \end{tabular}
  }
\end{table}

\paragraph{Setup.}
The primary comparison uses a Kimina-based stack to keep the core comparison controlled under the same generator-call budget.
To test whether the search layer still helps once the base stack is substantially stronger, we run a statement-generation evaluation on CombiBench under the same $T{=}100$ accounting with Goedel-Formalizer-V2-8B as the seeder and GPT-5.4 as the repair model.
We keep the semantic judge and statement-stage reporting protocol unchanged.

\paragraph{Compared methods.}
\textsc{Sample} uses Goedel-Formalizer-V2-8B alone, without repair or archive search.
\textsc{Compile+Semantic Repair} keeps the same stronger seeder and applies bounded compile / semantic repair without an archive.
\textsc{Hybrid control} keeps the same stronger seeder and stronger repair stack, with archive-based search disabled.
\textbf{\textsc{FormalEvolve}} then adds archive-based search on top of this no-archive control.

\paragraph{Interpretation.}
The stronger-base statement-generation setting preserves the main qualitative pattern.
Relative to the matched no-archive hybrid control, archive-based search still improves SH@100 (0.86 $\rightarrow$ 0.88), increases deduplicated SemOK$_\text{total}$ by 1.7$\times$ (1886 $\rightarrow$ 3238), and further reduces concentration (Gini 0.4832 $\rightarrow$ 0.4383; Top-10\% share 0.3128 $\rightarrow$ 0.2202).
The search layer both expands the total pool of judge-accepted candidates and makes that pool less concentrated.

\FloatBarrier

\subsection{What drives gains: repair versus online diversity}
Under fixed budgets, bounded patch/repair calls are the dominant contributor to semantic hit rate: removing patch repair reduces SH@100 by 11/100 (CombiBench) and 13/186 (ProofNet) (Table~\ref{tab:full-hit100}).
EvolAST-style fallback has a smaller and non-monotonic effect: relative to the no-EvolAST ablation, the run with EvolAST gains semantic hits on 11 CombiBench problems and loses them on 3, while on ProofNet it gains 1 and loses 5.
Because \texttt{semantic\_ok} is assigned by the LLM-as-judge implementation of $\SemOK$, these differences should be read as changes in the distribution of judge-accepted candidates rather than guaranteed semantic improvements.

\paragraph{Illustrative instances (EvolAST can help or hurt).}
The examples below show one instance where EvolAST enables a semantic hit and one where it suppresses a hit under the same $T{=}100$ budget.
\begin{center}
\small
\setlength{\tabcolsep}{5pt}
\resizebox{\columnwidth}{!}{%
\begin{tabular}{lcccc}
\toprule
Problem & Dataset & first hit call (w/ EvolAST / w/o) & semantic\_ok (w/ EvolAST / w/o) & $\Delta$ semantic\_ok \\
\midrule
\texttt{0011\_brualdi\_ch1\_10} & CombiBench & 22 / \texttt{no hit} & 12 / 0 & +12 \\
\texttt{0014\_brualdi\_ch2\_36} & CombiBench & \texttt{no hit} / 48 & 0 / 3 & -3 \\
\bottomrule
\end{tabular}
}
\end{center}
\FloatBarrier

\subsection{Failure modes}
We log (i) compilation failures, (ii) semantic drift (judge rejects), and (iii) judge/prover disagreements for qualitative inspection.
The qualitative cases in Appendix~\ref{sec:appendix-case-studies} show representative instances of these disagreements.

\paragraph{Example (judge/prover mismatch).}
In Appendix~\ref{sec:appendix-case-failure-baselinewin} (ProofNet \texttt{0004\_exercise\_6\_4\_2}), both baselines obtain theorem-complete prover outputs under the same RR64 budget, while our best judge-accepted statement is only pass-level.
This illustrates non-dominance between semantic judging and prover utility: semantic acceptance and proof success provide different evidence under fixed-budget evaluation.

\FloatBarrier

\section{Human and Prover-Stage Audits}
\label{sec:appendix-audits}

\subsection{Manual faithfulness audits}
\label{sec:appendix-manual-faithfulness-audit}

\paragraph{Audit overview.}
We report two complementary manual audits with the same \emph{Faithful / Partial / Severe} label set.
The first audit targets theorem-complete artifacts from the three original generator-based methods at the solved end of the pipeline.
The second targets the main statement-stage setting with a matched ProofNet audit over FormalEvolve, the hybrid no-archive control, and Sample.

\paragraph{Theorem-complete audit: scope and audit unit.}
We audit the theorem-complete end of the original three-method comparison pool.
The pool contains 161 theorem-complete artifacts across \textsc{Sample}, \textsc{C+S Repair}, and \textsc{FormalEvolve}: 29 from CombiBench and 132 from ProofNet.
The audit unit is one solved \emph{benchmark $\times$ method $\times$ problem} cell.
If a solved cell contains multiple theorem-complete attempts, we take the first theorem-complete attempt in \texttt{per\_problem.json} order and trace it back to the corresponding source statement in \texttt{rr64\_dataset.jsonl}.

\paragraph{Labels.}
\emph{Faithful} means that the proved formal statement matches the mathematical content of the informal target.
\emph{Partial} denotes a materially narrowed, incomplete, or otherwise only partially aligned statement.
\emph{Severe} denotes wrong-object, vacuous, or structurally broken cases that do not preserve the intended mathematical claim.

\begin{center}
  \captionof{table}{Pool-level totals for the manual faithfulness audit over theorem-complete artifacts in the original three-method comparison pool.}
  \label{tab:manual-faithfulness-pool}
  \scriptsize
  \begin{tabular}{lccc}
    \toprule
    Pool & Faithful & Partial & Severe \\
    \midrule
    CombiBench (29) & 18 & 4 & 7 \\
    ProofNet (132) & 93 & 13 & 26 \\
    Total (161) & 111 & 17 & 33 \\
    \bottomrule
  \end{tabular}
\end{center}

\begin{center}
  \captionof{table}{Method-level totals for the same manual faithfulness audit.}
  \label{tab:manual-faithfulness-method}
  \scriptsize
  \begin{tabular}{lccc}
    \toprule
    Method & Faithful & Partial & Severe \\
    \midrule
    \textsc{FormalEvolve} & 41 & 6 & 11 \\
    \textsc{Sample} & 34 & 5 & 10 \\
    \textsc{C+S Repair} & 36 & 6 & 12 \\
    \bottomrule
  \end{tabular}
\end{center}

\paragraph{Multi-reader cross-check.}
On a multi-reader cross-check subset drawn from the manually audited portion, the initial agreement rate on the \emph{Faithful / Partial / Severe} labels is 92\% before adjudication.
Disagreements were rechecked jointly, and uncertain cases were resolved conservatively.

\paragraph{Aggregate pattern.}
Within this audited theorem-complete pool, \textsc{FormalEvolve} yields the largest absolute count of \emph{Faithful} artifacts (41), while \emph{Severe} counts are numerically similar across methods (11, 10, and 12).
Proof success certifies the formal statement that was actually encoded and proved, but does not by itself establish faithfulness to the source statement; all three audited methods still contain severe mismatches at the solved end of the pipeline.
The audit therefore complements the main fixed-budget results with a direct theorem-faithfulness view at the solved end of the pipeline.

\paragraph{Matched statement-stage audit.}
Because the statement-stage semantic metrics rely on judge-positive outputs, we also audit matched judge-positive statement-stage outputs on ProofNet.
We sample 50 problems for which FormalEvolve, the hybrid no-archive control, and Sample each have a late-window compile+semantic judge-positive witness.
For each method and sampled problem, the audit unit is the latest deduplicated judge-positive candidate with insertion rank in $[20,100]$.
This design compares faithfulness profiles on a shared judge-positive slice; coverage is reported separately by SH@100.

\begin{center}
  \captionof{table}{Matched manual statement-stage audit on late-window judge-positive ProofNet outputs. The same 50 problems are audited for all three methods.}
  \label{tab:manual-faithfulness-statement}
  \scriptsize
  \begin{tabular}{lccc}
    \toprule
    Method & Faithful & Partial & Severe \\
    \midrule
    \textsc{FormalEvolve} & 42 & 5 & 3 \\
    Hybrid control & 43 & 4 & 3 \\
    Sample & 43 & 3 & 4 \\
    \bottomrule
  \end{tabular}
\end{center}
\FloatBarrier

\paragraph{Interpretation.}
Under final adjudication, the matched statement-stage audit shows similar faithfulness profiles across the three methods: FormalEvolve has 42/50 \emph{Faithful} candidates, Hybrid control has 43/50, and Sample has 43/50.
Severe counts are also close (3/50, 3/50, and 4/50 respectively).
In this slice, FormalEvolve's judge-positive outputs have a faithfulness profile similar to the two no-archive controls, with no obvious additional semantic shift from archive search.


\subsection{Prover-stage audit: what proof success certifies}
\label{sec:appendix-prover-audit}

\paragraph{Scope and claim boundary.}
We audit the prover stage on the subset of judge-accepted statements produced within the generator-call budget (\Cref{sec:budget}).
A prover-complete Lean file certifies that the produced formal artifact is accepted by Lean and contains no remaining \texttt{sorry}.
Faithfulness to the original informal statement is assessed by the LLM-as-judge implementation of $\SemOK$ and checked with manual audits (Appendix~\ref{sec:appendix-manual-faithfulness-audit}).

\paragraph{Prompt reference.}
The prover prompt template is fixed (Appendix~\ref{sec:appendix-prover-prompt}); the case studies report the proved formal statement and the prover response.

\paragraph{ASCII normalization.}
For compatibility with pdfLaTeX, raw transcripts in \texttt{PromptBox} are displayed after ASCII-only normalization of common Unicode math symbols (e.g., rendering quantifiers and connectives in ASCII form).
This affects presentation only.

\paragraph{Theorem-complete reporting.}
In addition to standard pass@64 and complete@64, we report \emph{theorem-complete@64}: a problem counts as theorem-complete@64 if at least one prover attempt yields a Lean-accepted, \texttt{sorry}-free file that contains an explicit \texttt{theorem} or \texttt{lemma}.
This guards against Lean-complete outputs that only introduce auxiliary \texttt{def}/\texttt{abbrev}/\texttt{example} content without proving a named proposition.
All prover-stage results below use the RR64 attempt schedule defined in \Cref{sec:appendix-metric-definitions}.

\paragraph{Prompt categories.}
Table~\ref{tab:appendix-prover-audit-stats} also reports a coarse prompt-type audit for the audited generator-based methods \emph{over prover attempts} (not over problems): the denominator is the number of prover prompts actually issued under a fixed per-problem attempt budget.
We label an attempt as ``Prompt has theorem'' if the candidate statement sent to the prover contains an explicit \texttt{theorem}/\texttt{lemma} declaration.
We label an attempt as ``Prompt abbrev-only'' if the candidate contains an \texttt{abbrev} declaration and no \texttt{theorem}/\texttt{lemma}.
Remaining attempts typically correspond to \texttt{def} or \texttt{example} templates and are omitted from this summary.

\begin{table*}[t]
  \centering
  \caption{Prover-stage audit statistics under $B{=}64$ prover attempts per problem for the audited generator-based methods. Prompt-category counts are reported \emph{per prover attempt}, not per problem. We report (i) whether the prover prompt contains a \texttt{theorem}/\texttt{lemma} declaration vs. abbrev-only prompts, and (ii) pass@64/complete@64 together with theorem-complete@64.}
  \label{tab:appendix-prover-audit-stats}
  \scriptsize
  \begin{tabular}{llccccc}
    \toprule
    Benchmark & Method & Prompt has theorem & Prompt abbrev-only & pass@64 & complete@64 & theorem-complete@64 \\
    \midrule
    CombiBench & FormalEvolve & 2299/3648 (0.630) & 384/3648 (0.105) & \PUCBPassOurs & \PUCBCompleteOurs & \PUCBTheoremCompleteOurs \\
    CombiBench & C+S Repair & 1813/2944 (0.616) & 384/2944 (0.130) & \PUCBPassStrongSemantic & \PUCBCompleteStrongSemantic & \PUCBTheoremCompleteStrongSemantic \\
    CombiBench & Sample & 1792/2816 (0.636) & 256/2816 (0.091) & \PUCBPassSample & \PUCBCompleteSample & \PUCBTheoremCompleteSample \\
    \midrule
    ProofNet & FormalEvolve & 9136/9536 (0.958) & 192/9536 (0.020) & \PUPNPassOurs & \PUPNCompleteOurs & \PUPNTheoremCompleteOurs \\
    ProofNet & C+S Repair & 8938/9280 (0.963) & 128/9280 (0.014) & \PUPNPassStrongSemantic & \PUPNCompleteStrongSemantic & \PUPNTheoremCompleteStrongSemantic \\
    ProofNet & Sample & 8164/8512 (0.959) & 128/8512 (0.015) & \PUPNPassSample & \PUPNCompleteSample & \PUPNTheoremCompleteSample \\
    \bottomrule
  \end{tabular}
\end{table*}
\FloatBarrier

\paragraph{Takeaway.}
On ProofNet, most prover prompts contain a \texttt{theorem}/\texttt{lemma} (around 0.96), so theorem-complete filtering is close to the standard metrics.
On CombiBench, 9--13\% of prompts are abbrev-only; theorem-complete@64 is therefore a stricter and more robust success signal.

\begin{figure*}[t]
  \centering
  \includegraphics[trim=6pt 6pt 6pt 6pt,clip,width=0.96\textwidth]{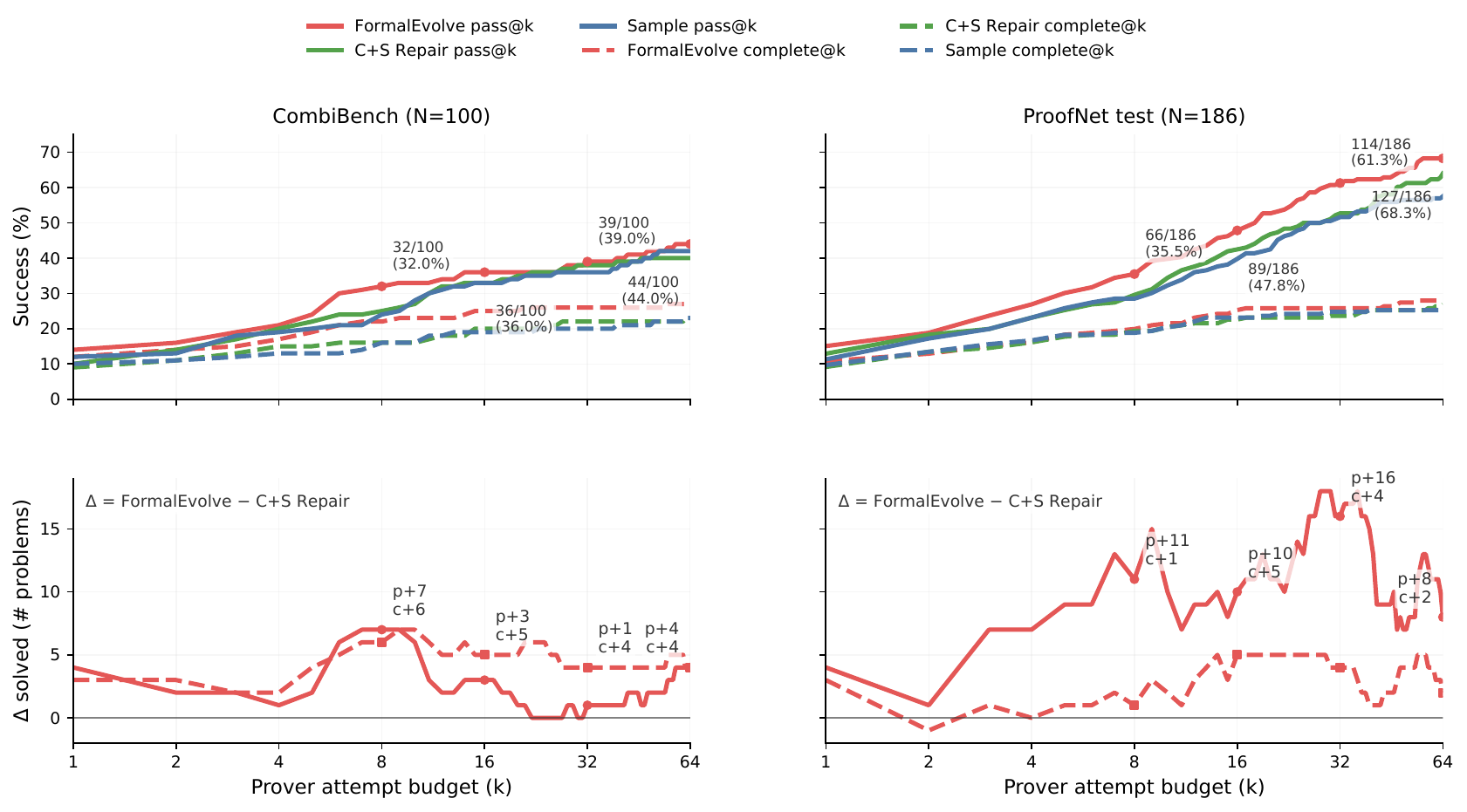}
  \caption{Proof utility as a function of prover attempts $k$ on the judge-accepted repertoire produced within $T=100$ calls. Solid: pass@k; dashed: complete@k; bottom row shows the gap relative to Kimina Compile+Semantic Repair.}
  \label{fig:proof-utility-panel-hitk}
\end{figure*}

\begin{figure*}[t]
  \centering
  \includegraphics[trim=6pt 6pt 6pt 6pt,clip,width=0.98\textwidth]{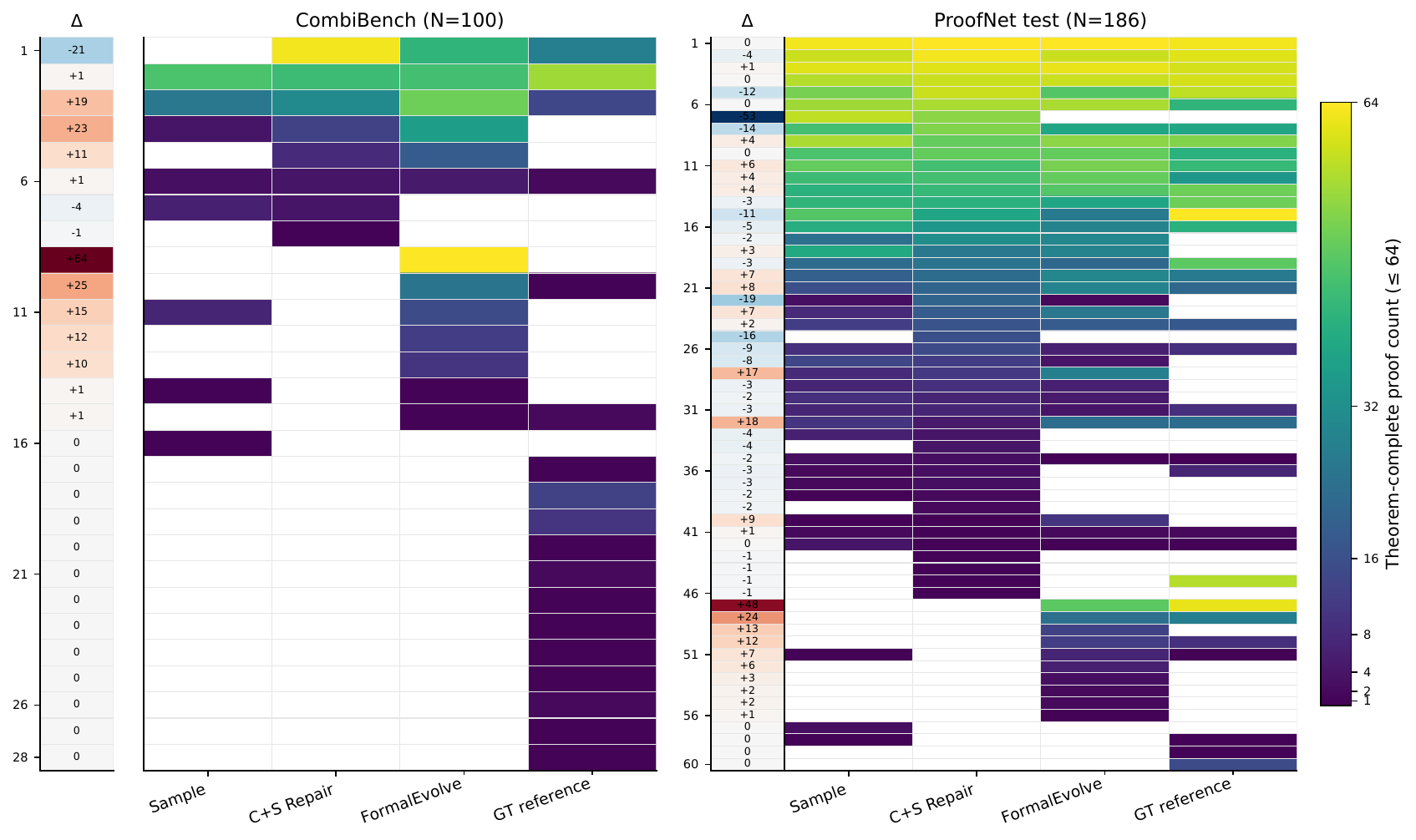}
  \caption{Filtered per-problem theorem-complete attempt counts under RR64 (cell = number of attempts, out of $B{=}64$, that yield a \texttt{theorem}/\texttt{lemma} without \texttt{sorry}). Rows show only problems where at least one method attains at least one theorem-complete attempt (so the number of displayed problems is smaller than $N$). Generator-based methods only attempt statements that pass the semantic judge (\texttt{semantic\_ok}$=1$), whereas the Ground truth column runs the prover on the dataset-provided ground-truth formal statement (one per problem), bypassing the judge.}
  \label{fig:theorem-complete-filtered-heatmap}
\end{figure*}

\begin{figure*}[t]
  \centering
  \includegraphics[trim=6pt 6pt 6pt 6pt,clip,width=0.98\textwidth]{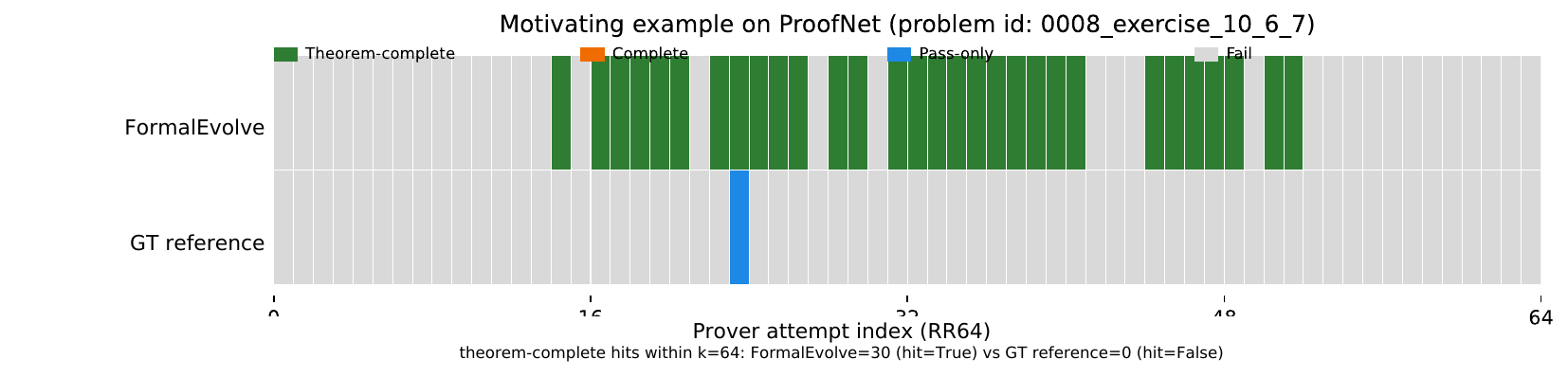}
  \caption{Data-driven motivating example under RR64 on ProofNet: even among judge-accepted candidates, prover outcomes can vary substantially under a fixed prover and attempt budget. The single-reference GT control uses one canonical statement (retried under RR64), while FormalEvolve provides a diverse judge-accepted repertoire, increasing the chance that at least one provable formulation is found within budget.}
  \label{fig:appendix-proof-utility-motivating-example}
\end{figure*}
\FloatBarrier

\FloatBarrier

\section{Case Studies and EvolAST Ablation}
\label{sec:appendix-case-studies}

We give representative statement/prover interactions and an EvolAST ablation to illustrate where archive search helps, where it is non-dominant, and how symbolic rewrites can change search trajectories under a fixed budget.

\subsection{Case studies: two successes and one non-dominance example}

\paragraph{Scope.}
We include two short success cases that illustrate prover-stage outcomes under the same RR64 budget (one where baselines fail to produce any Lean~4-accepted proof output, and one where a baseline reaches only pass-level artifacts), plus one counterexample that highlights non-dominance and judge/prover mismatch.
To contextualize these examples, Table~\ref{tab:appendix-sec5-nondom} quantifies prover-stage non-dominance frequency at theorem-complete@64 (FormalEvolve versus Compile+Semantic Repair) under the same RR64 protocol.

\subsubsection{Theorem-complete proof: FormalEvolve completes; baselines have no pass}
\label{sec:appendix-case-proof-ideal}
\noindent Benchmark: \texttt{proofnet}.\\
\noindent Problem id: \texttt{0088\_exercise\_4\_3\_1}.\\
\noindent Informal statement:
\begin{quote}\small
If $R$ is a commutative ring and $a \in R$, let $L(a) = \{x \in R \mid x a = 0\}$. Prove that $L(a)$ is an ideal of $R$.
\end{quote}

\paragraph{Outcome under the same fixed $B=64$ prover budget.}
Our run attains a theorem-complete proof, while both sampling baselines fail to produce any Lean~4-accepted proof output within $B=64$ attempts (no \texttt{pass}).

\paragraph{Takeaway.}
This instance illustrates a practical benefit of a repertoire: under a fixed prover attempt budget, the ability to try multiple judge-accepted statements can be the difference between reaching a complete proof and having no Lean-accepted proof output at all.

\paragraph{Notation.}
Lean~4 admits both Unicode and ASCII forms. For pdfLaTeX compatibility, we display prover transcripts in ASCII, where quantifiers/connectives may appear as \texttt{forall}/\texttt{->} and the constructor names \texttt{And}/\texttt{Or}, corresponding to $\forall/\to$ and $\wedge/\vee$.

\paragraph{Abridged prover output excerpt (FormalEvolve; theorem-complete).}
\begin{PromptBox}
import Mathlib

open Fintype Set Real Ideal Polynomial
open scoped BigOperators
noncomputable section

theorem my_theorem_123 {R : Type*} [CommRing R] (a : R) :
  forall x y : R, x * a = 0 -> y * a = 0 ->
    And ((x + y) * a = 0)
      (And ((-x) * a = 0)
        (forall r : R, And ((r * x) * a = 0) ((x * r) * a = 0))) := by
intro x y hx hy
have h1 : (x + y) * a = 0 := by
  calc (x + y) * a = x * a + y * a := by ring
       _ = 0 := by simp [hx, hy]
have h2 : (-x) * a = 0 := by
  calc (-x) * a = - (x * a) := by ring
       _ = 0 := by simp [hx]
have h3 : forall r : R, And ((r * x) * a = 0) ((x * r) * a = 0) := by
  intro r
  have h4 : (r * x) * a = 0 := by calc (r * x) * a = r * (x * a) := by ring; _ = 0 := by simp [hx]
  have h5 : (x * r) * a = 0 := by calc (x * r) * a = (r * x) * a := by ring; _ = 0 := h4
  exact And.intro h4 h5
exact And.intro h1 (And.intro h2 h3)
\end{PromptBox}
\begin{PromptBox}
FormalEvolve: pass=True,  complete=True,  sorries=0
C+S Repair:  pass=False
Sample:  pass=False
\end{PromptBox}

\subsubsection{Complete proof: FormalEvolve completes; baseline is pass-only}
\label{sec:appendix-case-proof-completion}
\noindent Benchmark: \texttt{proofnet}.\\
\noindent Problem id: \texttt{0000\_exercise\_2\_3\_2}.\\
\noindent Informal statement:
\begin{quote}\small
Prove that the products $a b$ and $b a$ are conjugate elements in a group.
\end{quote}

\paragraph{Outcome under the same fixed $B=64$ prover budget.}
Our run reaches a theorem-complete proof (\texttt{complete=True, sorries=0}), while Compile+Semantic Repair only reaches pass-level artifacts with remaining \texttt{sorry}.

\paragraph{Prover output excerpt (FormalEvolve; theorem-complete).}
\begin{PromptBox}
import Mathlib

theorem my_theorem_xxx {G : Type*} [Group G] (a b : G) :
  exists g : G, a * b = g * (b * a) * g^(-1) := by
  refine Exists.intro a ?_
  simp [mul_assoc, mul_left_inv, mul_right_inv]
\end{PromptBox}

\begin{figure}[t]
  \centering
  \includegraphics[trim=6pt 6pt 6pt 6pt,clip,width=\columnwidth]{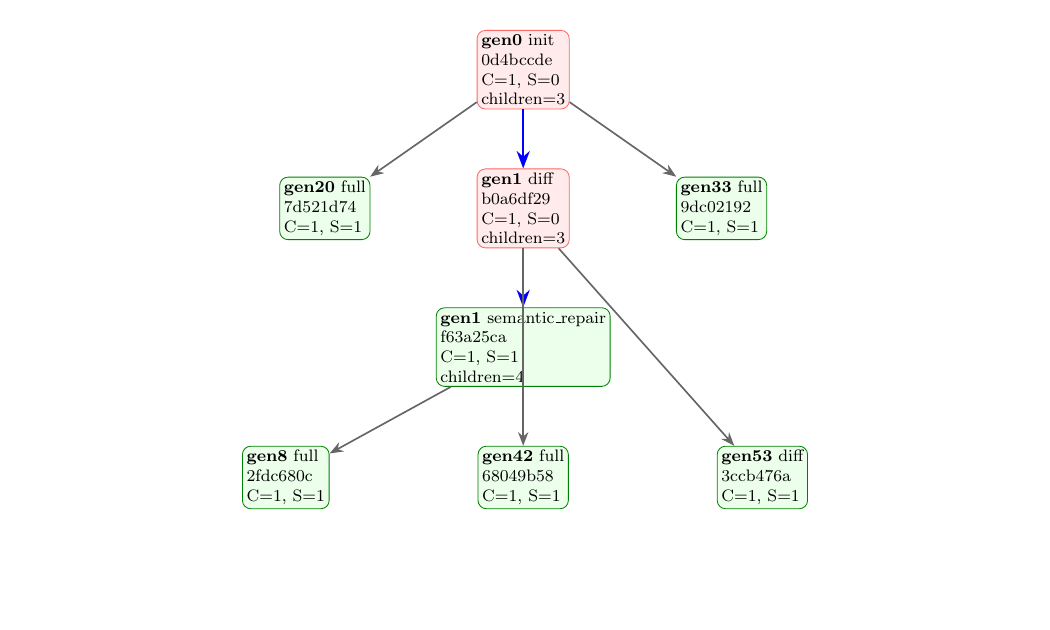}
  \caption{ProofNet \texttt{0000\_exercise\_2\_3\_2} statement evolution trace (Comp=compile-ok, Sem=semantic-ok). The repair step removes an extra conjunction and introduces the conjugation structure (the $g^{-1}$ term).}
  \label{fig:pn0000_evo_trace}
\end{figure}

\paragraph{Statement evolution trace (mechanism note).}
This instance illustrates why we separate compilation from semantic checking: an early candidate may compile yet express a non-standard (and generally stronger) variant of the target claim, which can be judged inconsistent and/or harder for the prover to use under a fixed attempt budget.
Here, the parent uses a conjunction of two equations, whereas the informal goal is a single conjugacy statement of the form $a*b = g*(b*a)*g^{-1}$.
Bounded semantic repair removes the extra conjunct and rewrites the statement into this standard form while preserving compilation feasibility (Figure~\ref{fig:pn0000_evo_trace}).

\begin{PromptBox}
init (Comp=1, Sem=0): exists g, a*b = g*b*a /\ b*a = g*a*b
repair (Comp=1, Sem=1): exists g, a*b = g*(b*a)*g^(-1)
\end{PromptBox}

\subsubsection{Non-dominance case: baseline theorem-complete; FormalEvolve pass-only}
\label{sec:appendix-case-failure-baselinewin}
\noindent Benchmark: \texttt{proofnet}.\\
\noindent Problem id: \texttt{0004\_exercise\_6\_4\_2}.\\
\noindent Informal statement:
\begin{quote}\small
Prove that no group of order p q, where p and q are prime, is simple.
\end{quote}

\paragraph{Outcomes under the same fixed $B=64$ prover budget.}
In this instance, our best judge-accepted statement yields \texttt{pass=True} but not a complete proof, while both sampling baselines obtain a theorem-complete proof within the same prover attempt budget.

\paragraph{Takeaway.}
This example highlights non-dominance: even under the same prover budget, different judge-accepted statements can differ substantially in proof-search friendliness, so proof utility and semantic faithfulness need separate readouts.

\paragraph{Prover output excerpt (FormalEvolve; pass-only).}
The excerpt below contains remaining \texttt{sorry}; the baseline theorem-complete outputs serve as the comparison point through the reported outcome flags.
\noindent FormalEvolve (pass-only):\\
\begin{PromptBox}
import Mathlib

theorem my_theorem_12345 {G : Type*} [Group G] [Fintype G] (p q : Nat)
  (hp : p.Prime) (hq : q.Prime) (hcard : Fintype.card G = p * q) :
    Not (IsSimpleGroup G) := by
  have h_main : Not (IsSimpleGroup G) := by sorry
  sorry
\end{PromptBox}
\begin{PromptBox}
FormalEvolve: pass=True,  complete=False, sorries=1
C+S Repair:  pass=True,  complete=True,  sorries=0
Sample:  pass=True,  complete=True,  sorries=0
\end{PromptBox}
\FloatBarrier

\subsection{Early proof successes under fixed prover budgets}
\label{sec:appendix-early-proof-wins}
Table~\ref{tab:early-proof-wins} lists representative instances where FormalEvolve reaches a successful proof substantially earlier than Compile+Semantic Repair under the same prover attempt budget, or succeeds when the baseline fails within $B=64$ attempts.
We report the first prover attempt index $k$ (under the round-robin schedule described in \Cref{sec:appendix-metric-definitions}) where the prover returns (i) any proof script accepted by Lean~4 (pass) or (ii) a complete proof without \texttt{sorry} (complete).

\begin{table*}[t]
  \centering
  \caption{Representative early proof successes under a fixed per-problem prover attempt budget $B=64$, computed on the judge-accepted repertoire produced within $T=100$ generator calls. We report the first prover attempt index $k$ where the prover succeeds; ``$>64$'' indicates no success within the attempt budget.}
  \label{tab:early-proof-wins}
  \scriptsize
  \setlength{\tabcolsep}{2pt}
  \begin{tabular}{@{}>{\raggedright\arraybackslash}p{0.12\textwidth}>{\raggedright\arraybackslash}p{0.09\textwidth}>{\raggedright\arraybackslash}p{0.50\textwidth}>{\centering\arraybackslash}p{0.13\textwidth}>{\centering\arraybackslash}p{0.12\textwidth}@{}}
    \toprule
    Benchmark & Criterion & Problem (id + short description) & FormalEvolve first $k$ & C+S Repair first $k$ \\
    \midrule
    \multicolumn{5}{l}{\textbf{Both succeed, but FormalEvolve is much earlier}} \\
    ProofNet & pass & \path{0113_exercise_18_8a} -- order topology: $\{x\mid f(x)\le g(x)\}$ is closed & 1 & 58 \\
    ProofNet & complete & \path{0113_exercise_18_8a} -- order topology: $\{x\mid f(x)\le g(x)\}$ is closed & 12 & 58 \\
    ProofNet & pass & \path{0174_exercise_5_3} -- $f(x)=x+\varepsilon g(x)$ is injective for small $\varepsilon$ & 2 & 64 \\
    ProofNet & complete & \path{0174_exercise_5_3} -- $f(x)=x+\varepsilon g(x)$ is injective for small $\varepsilon$ & 13 & 64 \\
    CombiBench & pass & \path{0047_brualdi_ch13_10} -- tournament radius-2 vertex exists & 9 & 27 \\
    CombiBench & complete & \path{0047_brualdi_ch13_10} -- tournament radius-2 vertex exists & 21 & 27 \\
    CombiBench & pass & \path{0057_imosl_2015_c6} -- infinitely many non-clean integers & 1 & 5 \\
    CombiBench & complete & \path{0057_imosl_2015_c6} -- infinitely many non-clean integers & 3 & 15 \\
    \midrule
    \multicolumn{5}{l}{\textbf{FormalEvolve succeeds, baseline fails within $B=64$}} \\
    ProofNet & complete & \path{0000_exercise_2_3_2} -- $ab$ and $ba$ are conjugate in a group & 2 & $>64$ \\
    ProofNet & complete & \path{0046_exercise_3_4_5b} -- solvable quotients are solvable & 1 & $>64$ \\
    CombiBench & complete & \path{0007_hackmath_8} -- ferry crossing: women in first group & 1 & $>64$ \\
    CombiBench & complete & \path{0063_usamo_2000_p4} -- 3 colored squares make a right triangle & 1 & $>64$ \\
    \bottomrule
  \end{tabular}
\end{table*}
\FloatBarrier

\subsection{Ablation note: when EvolAST changes outcomes on ProofNet}
\label{sec:appendix-evolast-proofnet}
On ProofNet, enabling EvolAST slightly decreases semantic hit@100 under fixed-budget accounting (158/186 vs 162/186).
The gap is small (4 problems), but it illustrates an important point for budgeted search: a diversity operator can change the search trajectory and archive composition, and may not be universally beneficial across datasets.

\paragraph{Where the difference comes from.}
Under fixed-budget accounting, there are 5 problems that achieve at least one judge-accepted statement under the no-EvolAST configuration but not under the main configuration, and 1 problem where the reverse holds.
Because the generator and patching process are stochastic, these disagreements reflect diverging search trajectories under nearly identical protocols; attributing causality to any single operator requires careful per-instance inspection.
Empirically, these disagreements are consistent with a mixture of direct trigger effects (duplicates/compile-fail fallbacks), indirect trajectory effects, and dataset-dependent operator behavior.

\begin{table*}[t]
  \centering
  \caption{ProofNet instances where the main run and the no-EvolAST ablation disagree on semantic hit@100 under fixed-budget accounting. We report the first call index where a semantic success appears (when it does), computed from the budget accounting logs.}
  \label{tab:evolast-proofnet-disagreements}
  \scriptsize
  \setlength{\tabcolsep}{2pt}
  \begin{tabular}{p{0.18\textwidth}p{0.42\textwidth}>{\centering\arraybackslash}p{0.17\textwidth}>{\centering\arraybackslash}p{0.17\textwidth}}
    \toprule
    Problem id & Short description & Main (repair + EvolAST) & No EvolAST (repair only) \\
    \midrule
    \path{0066_exercise_8_3_6b} & $\mathbb{Z}[i]/(q)$ is a field with $q^2$ elements (for $q\equiv 3 \bmod 4$) & no hit & hit at call 42 \\
    \path{0070_exercise_9_4_9} & $x^2-\sqrt{2}$ irreducible over $\mathbb{Z}[\sqrt{2}]$ & no hit & hit at call 65 \\
    \path{0142_exercise_4_15a} & Uniform continuity iff having a modulus of continuity & no hit & hit at call 49 \\
    \path{0151_exercise_1_8} & No ordered-field structure on $\mathbb{C}$ & no hit & hit at call 39 \\
    \path{0185_exercise_5_1} & Blaschke condition: $\sum_n (1-|z_n|)<\infty$ for bounded holomorphic $f$ & no hit & hit at call 66 \\
    \midrule
    \path{0123_exercise_25_9} & Identity component is a normal subgroup & hit at call 87 & no hit \\
    \bottomrule
  \end{tabular}
\end{table*}

\paragraph{Representative example (problem \texttt{0142\_exercise\_4\_15a}).}
In this instance, the main configuration produced many compilable candidates but no semantic success, whereas the no-EvolAST configuration eventually reaches a judge-accepted statement after semantic repair.
One characteristic we observe in the main run is that EvolAST fallback triggers repeatedly on duplicates and compile failures, producing compilable but judge-rejected variants; in this problem, many candidates are explicitly tagged as EvolAST fallbacks, and all have $\CompOK(c)=1$ but $\SemOK(c)=0$.
This can bias the archive toward structurally perturbed but semantically misaligned candidates under a fixed budget, which in turn alters parent selection and patch contexts.

Concretely, the semantic judge flags a common failure mode here: \emph{domain mismatch} (e.g., proving uniform continuity on $\mathbb{R}$ with the intended $[a,b]$ restriction missing).
The best candidate found by the main run (compilable but judge-rejected) has the unconstrained target:
\begin{quote}\small
\resizebox{\linewidth}{!}{$\left(\exists \mu,\ \cdots\ \wedge\ \forall s,t\in \mathbb{R},\ |f(s)-f(t)| \le \mu(|s-t|)\right)\ \leftrightarrow\ \texttt{UniformContinuous}\ f$}
\end{quote}
whereas the successful no-EvolAST semantic-repair output explicitly restricts the bound to $s,t\in [a,b]$ and targets uniform continuity on the interval:
\begin{quote}\small
\resizebox{\linewidth}{!}{$\left(\exists \mu,\ \cdots\ \wedge\ \forall s,t,\ s\in \texttt{Icc}\ a\ b \rightarrow t\in \texttt{Icc}\ a\ b \rightarrow |f(s)-f(t)| \le \mu(|s-t|)\right)$}\\
\resizebox{\linewidth}{!}{$\leftrightarrow\ \texttt{UniformContinuousOn}\ f\ (\texttt{Icc}\ a\ b)$}
\end{quote}
This illustrates a fundamental limitation of a conservative type-rewrite operator: our EvolAST rule set is limited to the existing binder and goal AST, so new problem-specific binders (e.g., explicit $a,b$ interval parameters) and new domain-restriction hypotheses require LLM repair.
Therefore, when the current parent trajectory has not yet reached a semantically correct ``skeleton'', applying EvolAST tends to preserve the same semantic mistake, producing many feasible variants that remain judge-rejected.
This effect is amplified by our implementation choice that EvolAST candidates terminate the fallback path to keep accounting bounded, leaving later semantic correction to subsequent LLM proposals.

\paragraph{Takeaway.}
Online diversity operators can have dataset-dependent effects on semantic hit rates under a fixed budget.
In practice, this motivates tunable EvolAST rule sets, tunable trigger conditions, and transparent ablation reporting.

\paragraph{EvolAST rewrite surface.}
In our experiments, EvolAST is used in its rule-based mode: it applies a bounded sequence of conservative, semantics-intended AST rewrites to the theorem type (binder types and goal type), keeping the proof body unchanged.
The rewrite surface is intentionally controlled and includes conservative hypothesis reordering; commutativity/associativity/distributivity for conjunction/disjunction (Lean: $\wedge/\vee$); symmetry swaps (e.g., $a=b$ to $b=a$, $P\leftrightarrow Q$ to $Q\leftrightarrow P$); and dual relations (e.g., $a<b$ to $b>a$, $a\le b$ to $b\ge a$).
\FloatBarrier

\FloatBarrier

\section{Algorithm, Budget Accounting, and Configuration}
\label{sec:appendix-reproducibility-budget}

We give the detailed per-problem search procedure, generator-call accounting rules, semantic-judge overhead, and implementation configuration used in the fixed-budget experiments.

\subsection{Detailed Pseudocode}
\label{sec:appendix-pseudocode}

\paragraph{Accounting conventions.}
Under fixed-budget accounting, we debit only generator-side LLM calls (proposals and bounded repairs).
EvolAST is a symbolic fallback triggered on (i) exact duplicates after canonicalization and (ii) compilation failures; it does not consume generator-call budget and is evaluated as the representative candidate of the triggering call.
Each compilation-repair or semantic-repair attempt is itself a debited generator call whose representative candidate is the repaired output; semantic repair preserves the original compilable candidate and appends a repaired variant, and both are evaluated.
All metrics normalize by the fixed budget $T$ per problem; any early termination or failed call is treated as a consumed call with $\CompOK(c)=0$ and $\SemOK(c)=0$.

\paragraph{Parent selection.}
Parent selection uses the usage-penalized robust weighting described in Section~\ref{sec:method}.
For completeness, each archived candidate $c_i$ has gate score $s_i=\CompOK(c_i)(1+\SemOK(c_i))$ and parent-usage count $n_i$.
Within the selected island, we set $z_i=(s_i-m)/d$, where $m=\mathrm{median}_j(s_j)$ and $d=\max(\mathrm{MAD}(\{s_j\}_j),10^{-6})$.
The quality factor is $Q_i=\sigma(\lambda z_i)$, the reuse factor is $R_i=[1+(1+\beta)n_i]^{-1}$, and the next parent is sampled with probability $Q_iR_i/\sum_j Q_jR_j$.
Here $\mathrm{MAD}(A)=\mathrm{median}_{a\in A}|a-\mathrm{median}(A)|$, $10^{-6}$ is a numerical stabilizer, and $R_i$ discourages repeatedly selecting the same parent.

\begin{algorithm}[H]
  \caption{FormalEvolve (per problem; detailed)}
  \label{alg:formalevolve-detail}
  \begin{algorithmic}
    \footnotesize
    \STATE {\bfseries Input:} informal statement $x$, generator-call budget $T$, seed model $M_{\mathrm{seed}}$, patch model $M_{\mathrm{patch}}$, predicates $\CompOK,\SemOK$, islands $K$
    \STATE {\bfseries Output:} feasible archive $\cup_I\mathcal{A}_I$ partitioned into islands
    \STATE $t\leftarrow 0$; $\mathcal{A}_{1:K}\leftarrow (\emptyset,\ldots,\emptyset)$
    \STATE $(\mathcal{A}_{1:K},t)\leftarrow \textsc{InitSeeds}(x,M_{\mathrm{seed}},T,K;\CompOK,\SemOK)$
    \STATE \hfill{\footnotesize (debits budget; compiles/judges seeds; inserts feasible seeds into islands)}
    \WHILE{$t < T$ and $\bigcup_I\mathcal{A}_I\neq\emptyset$}
      \STATE $I \leftarrow \textsc{SampleIsland}(\{I:\mathcal{A}_I\neq\emptyset\})$ \hfill (uniform over nonempty islands)
      \STATE $p \leftarrow \textsc{SampleParent}(\mathcal{A}_I)$; $n_p\leftarrow n_p+1$
      \STATE $(\mathcal{I}_{\mathrm{arch}},\mathcal{I}_{\mathrm{top}})\leftarrow \textsc{SampleContext}(\mathcal{A}_I,p)$
      \STATE $c \leftarrow \textsc{Propose}(p,\mathcal{I}_{\mathrm{arch}},\mathcal{I}_{\mathrm{top}},x;M_{\mathrm{patch}})$
      \STATE $t\leftarrow t+1$
      \STATE $c \leftarrow \textsc{Enforce}(c,p)$ \hfill (preamble/proof protocol)
      \IF{\textsc{IsExactDuplicate}$(c,\mathcal{A}_I)$}
        \STATE $c_{\mathrm{rep}} \leftarrow \textsc{EvolAST}(p)$ \hfill (generator-call-free representative)
        \STATE \textsc{ProcessCandidate}$(c_{\mathrm{rep}},x,p,\mathcal{A}_I,t,\mathrm{false})$ \hfill (no semantic repair)
      \ELSIF{\textsc{Compiles}$(c)$}
        \STATE \textsc{ProcessCandidate}$(c,x,p,\mathcal{A}_I,t,\mathrm{true})$
      \ELSE
        \STATE $c_{\mathrm{rep}} \leftarrow \textsc{EvolAST}(p)$ \hfill (compile-fail representative; no call)
        \STATE \textsc{ProcessCandidate}$(c_{\mathrm{rep}},x,p,\mathcal{A}_I,t,\mathrm{false})$
        \IF{$t < T$}
          \STATE $c_{\mathrm{fix}} \leftarrow \textsc{CRepair}(c,x;M_{\mathrm{patch}})$; $t\leftarrow t+1$
          \STATE $c_{\mathrm{fix}} \leftarrow \textsc{Enforce}(c_{\mathrm{fix}},p)$
          \STATE \textsc{ProcessCandidate}$(c_{\mathrm{fix}},x,p,\mathcal{A}_I,t,\mathrm{true})$
        \ENDIF
      \ENDIF
      \STATE \textsc{MaybeMigrate}$(\mathcal{A}_1,\ldots,\mathcal{A}_K)$ \hfill (periodic island migration; see \Cref{sec:appendix-config})
    \ENDWHILE
  \end{algorithmic}
\end{algorithm}

\begin{algorithm}[H]
  \caption{\textsc{ProcessCandidate} (compile gate, judge, and bounded semantic repair)}
  \label{alg:appendix-process-candidate}
  \begin{algorithmic}
    \footnotesize
    \STATE {\bfseries Input:} candidate $c$, informal statement $x$, parent $p$, current island archive $\mathcal{A}_I$, call counter $t$, allow semantic repair flag $r\in\{\mathrm{true},\mathrm{false}\}$
    \STATE {\bfseries Effect:} if $c$ is feasible, evaluates $\SemOK(c)$ and inserts into $\mathcal{A}_I$; if $\SemOK(c)=0$ and $r=\mathrm{true}$, spends one additional generator call on bounded semantic repair and evaluates the repaired candidate
    \IF{\textsc{Compiles}$(c)$}
      \STATE $m \leftarrow \textsc{Evaluate}(c,x)$ \hfill (semantic judge)
      \STATE $\mathcal{A}_I\leftarrow \textsc{Insert}(\mathcal{A}_I,(c,m))$ \hfill ($\CompOK(c)=1$ only)
      \IF{$r \land \SemOK(c)=0 \land t < T$}
        \STATE $c' \leftarrow \textsc{SRepair}(c,x;M_{\mathrm{patch}})$; $t\leftarrow t+1$
        \STATE $c' \leftarrow \textsc{Enforce}(c',p)$
        \IF{\textsc{Compiles}$(c')$}
          \STATE $m' \leftarrow \textsc{Evaluate}(c',x)$
          \STATE $\mathcal{A}_I\leftarrow \textsc{Insert}(\mathcal{A}_I,(c',m'))$
        \ENDIF
      \ENDIF
    \ENDIF
  \end{algorithmic}
\end{algorithm}
\FloatBarrier

\subsection{Budget Audit and Evaluation Overhead}
\label{sec:appendix-budget-audit}

\paragraph{Generator-side.}
We audit generator-call usage under the budget currency ($T=100$ calls per problem).
We decompose the fixed budget $N\times T$ into generation calls (Gen), compilation-repair calls (CRep), and semantic-repair calls (SRep).
Table~\ref{tab:appendix-sec5-call-comp} provides a complementary per-problem distributional view (median/IQR) of CRep/SRep under the same accounting.

\paragraph{Evaluator-side (semantic judge).}
We report evaluator overhead in terms of semantic LLM-as-judge calls (CriticLean-Qwen3-14B).
In this pipeline, semantic judging is invoked only after successful compilation, and EvolAST outputs are judged under the same gate; EvolAST-Judge is the subset of judge calls attributed to EvolAST-labeled candidates (generator-call-free on the generator side).

\begin{table*}[t]
  \centering
  \caption{Strict generator-call budget audit (call currency). Budget is fixed to $N\times T$ with $T=100$ per problem; Gen/CRep/SRep partition the same budget.}
  \label{tab:appendix-budget-audit-generator}
  \scriptsize
  \setlength{\tabcolsep}{4pt}
  \begin{tabular}{llrrrr}
    \toprule
    Dataset & Method & Budget & Gen & CRep & SRep \\
    \midrule
    CombiBench & Sample & 10000 & 10000 & 0 & 0 \\
    CombiBench & Compile Repair & 10000 & 7031 & 2969 & 0 \\
    CombiBench & C+S Repair & 10000 & 4257 & 1991 & 3752 \\
    CombiBench & Hybrid control & 10000 & 4081 & 1855 & 4064 \\
    CombiBench & FormalEvolve & 10000 & 4895 & 3042 & 2063 \\
    CombiBench & FormalEvolve w/o EvolAST & 10000 & 5139 & 3428 & 1433 \\
    CombiBench & FormalEvolve w/o patch repair & 10000 & 10000 & 0 & 0 \\
    \midrule
    ProofNet & Sample & 18600 & 18600 & 0 & 0 \\
    ProofNet & Compile Repair & 18600 & 13881 & 4719 & 0 \\
    ProofNet & C+S Repair & 18600 & 10732 & 3987 & 3881 \\
    ProofNet & Hybrid control & 18600 & 10887 & 3700 & 4013 \\
    ProofNet & FormalEvolve & 18600 & 11872 & 4903 & 1825 \\
    ProofNet & FormalEvolve w/o EvolAST & 18600 & 11984 & 4887 & 1729 \\
    ProofNet & FormalEvolve w/o patch repair & 18600 & 18600 & 0 & 0 \\
    \bottomrule
  \end{tabular}
\end{table*}

\begin{table*}[t]
  \centering
  \caption{Semantic judge overhead (CriticLean-Qwen3-14B). JudgeCalls counts how many compilable candidates reach semantic checking; EvolAST-Judge is the EvolAST-attributed subset and is included in JudgeCalls.}
  \label{tab:appendix-budget-audit-judge}
  \scriptsize
  \setlength{\tabcolsep}{4pt}
  \begin{tabular}{llrrr}
    \toprule
    Dataset & Method & Problems & JudgeCalls & EvolAST-Judge \\
    \midrule
    CombiBench & Sample & 100 & 6953 & 0 \\
    CombiBench & Compile Repair & 100 & 5513 & 0 \\
    CombiBench & C+S Repair & 100 & 6109 & 0 \\
    CombiBench & FormalEvolve & 100 & 4970 & 863 \\
    CombiBench & FormalEvolve w/o EvolAST & 100 & 3800 & 0 \\
    CombiBench & FormalEvolve w/o patch repair & 100 & 9025 & 3797 \\
    \midrule
    ProofNet & Sample & 186 & 13062 & 0 \\
    ProofNet & Compile Repair & 186 & 11624 & 0 \\
    ProofNet & C+S Repair & 186 & 11904 & 0 \\
    ProofNet & FormalEvolve & 186 & 10536 & 331 \\
    ProofNet & FormalEvolve w/o EvolAST & 186 & 10103 & 0 \\
    ProofNet & FormalEvolve w/o patch repair & 186 & 13465 & 1008 \\
    \bottomrule
  \end{tabular}
\end{table*}

\begin{table*}[t]
  \centering
  \caption{Complete test-time accounting for the methods in Figure~\ref{fig:cost-performance}. All methods use $T{=}100$ generator calls per problem. G/C/S partitions generation, compilation-repair, and semantic-repair calls; Judge [E] gives total semantic-judge calls with the EvolAST-attributed subset in brackets; TC@64 is theorem-complete@64. Batch h reports the recorded end-to-end batch duration.}
  \label{tab:appendix-complete-cost-accounting}
  \scriptsize
  \setlength{\tabcolsep}{3.5pt}
  \resizebox{\textwidth}{!}{%
  \begin{tabular}{llrrrrrrr}
    \toprule
    Dataset & Method & SH@100 & TC@64 & G/C/S & Judge [E] & Lean eval. & Records & Batch h \\
    \midrule
    CombiBench & Sample & 0.440 & 8/100 & 10,000/0/0 & 6,953 [0] & 10,000 & 10,000 & 6.57 \\
    CombiBench & C+S Repair & 0.460 & 8/100 & 4,257/1,991/3,752 & 6,109 [0] & 10,000 & 7,790 & 7.22 \\
    CombiBench & Hybrid control & 0.530 & 9/100 & 4,081/1,855/4,064 & 3,881 [0] & 6,679 & 6,701 & 8.18 \\
    CombiBench & FormalEvolve & 0.580 & 13/100 & 4,895/3,042/2,063 & 4,970 [863] & 8,891 & 9,011 & 8.16 \\
    \midrule
    ProofNet & Sample & 0.715 & 41/186 & 18,600/0/0 & 13,062 [0] & 18,600 & 18,600 & 15.53 \\
    ProofNet & C+S Repair & 0.780 & 46/186 & 10,732/3,987/3,881 & 11,904 [0] & 18,600 & 14,684 & 18.25 \\
    ProofNet & Hybrid control & 0.828 & 40/186 & 10,887/3,700/4,013 & 10,565 [0] & 14,167 & 14,538 & 20.77 \\
    ProofNet & FormalEvolve & 0.849 & 45/186 & 11,872/4,903/1,825 & 10,536 [331] & 14,376 & 14,568 & 17.29 \\
    \bottomrule
  \end{tabular}%
  }
\end{table*}
\FloatBarrier

\subsection{Configuration and hyperparameters}
\label{sec:appendix-config}

Table~\ref{tab:appendix-config-combibench} summarizes the key hyperparameters used in the main experiments.

\paragraph{Lean~4 toolchain and library.}
All candidates are compiled under Lean~4.15.0 with Mathlib (\texttt{import Mathlib}).
We use Mathlib4 \texttt{v4.15.0} (commit \texttt{9837ca9}) as bundled by our Kimina Lean Server installation.
Compilation is served by Kimina Lean Server \citep{dossantos2025kimina}, which provides a pinned Lean~4/Mathlib environment suitable for large-scale batch checking.

\begin{table*}[t]
  \centering
  \caption{Key hyperparameters used in FormalEvolve.}
  \label{tab:appendix-config-combibench}
  \small
  \begin{tabular}{lp{0.72\textwidth}}
    \toprule
    Component & Setting \\
    \midrule
    Models & Seed: Kimina-Autoformalizer-7B; patch/repair: Qwen3-30B-A3B. \\
    Compilation & Lean~4.15.0 + Mathlib via Kimina Lean Server \citep{dossantos2025kimina}. \\
    Semantic judge & CriticLean-Qwen3-14B (LLM-based judge; proof ignored). \\
    Islands & 2 \\
    Island separation & enabled; parent selection and inspirations restricted to the sampled island \\
    Archive invariant & compilation-feasible (compile\_ok$=1$) \\
    Archive size & global capacity 40 (compilation-feasible) \\
    Migration policy & interval 10 gens, rate 0.1; moves feasible gen$>0$ candidates; elitism (top-1 protected) \\
    Parent selection & weighted ($\lambda=10$) with usage penalty $\beta=0.05$ \\
    Operator mix (default) & \texttt{full}/\texttt{diff}/\texttt{cross} with probs 0.5/0.3/0.2 \\
    Patch attempts & max patch attempts 1 (per step) \\
    Inspiration pool & per step: 4 archive inspirations + 2 top-$k$ inspirations (when available) \\
    Cross prompt inspirations & samples $k=1$ inspirations uniformly from the pool \\
    Repair budget & max attempts 2, temperature 0.7 (applies to compile + semantic repair) \\
    EvolAST fallback & enabled as a fallback for (i) exact-duplicate edits, and (ii) compilation failures (evaluated as a generator-call-free fallback alongside bounded compile repair) \\
    \bottomrule
  \end{tabular}
\end{table*}

\paragraph{Implementation notes: archive and island mechanics.}
\begin{itemize}
  \item \textbf{Global feasible archive.} We maintain a global compilation-feasible archive (capacity 40); selection is applied only within this feasible set.
  \item \textbf{Island-local behavior.} Parent selection and cross-patch inspirations are restricted to the sampled island (via \texttt{island\_idx}); children inherit the parent island by default.
  \item \textbf{Migration.} Every 10 generations, we migrate a small fraction of feasible candidates between islands (rate 0.1), while protecting the top-1 elite in each island.
\end{itemize}
\FloatBarrier

\FloatBarrier


\section{Prompting and Semantic-Judge Protocol}
\label{sec:appendix-prompts}

We report the prompt templates used for initial statement generation, evolution operators, bounded repair, semantic judging, and downstream proof completion.
We show the full prompt skeletons because prompt details affect both reproducibility and the reliability of LLM-as-judge semantic labels.

\paragraph{Notation.}
Placeholders include \{informal\} (natural-language statement), \{code\_content\} (current Lean file), \{performance\_metrics\} (compile/semantic metrics), \{text\_feedback\_section\} (optional judge feedback), \{original\_code\} (candidate file for repair), \{compile\_error\_type\}/\{compile\_error\_msg\} (compiler feedback), \{critic\_feedback\} (semantic-judge feedback), \{inspiration\_code\_1\} (cross inspiration), and \{lean\_statement\} (theorem statement for judging).
For pdfLaTeX compatibility, we render common Lean unicode symbols in ASCII in verbatim displays.

\begin{table*}[t]
  \centering
  \caption{Prompt components used in our pipeline.}
  \label{tab:appendix-prompts-summary}
  \small
  \begin{tabular}{@{}>{\raggedright\arraybackslash}p{0.22\textwidth}@{\hspace{8pt}}>{\raggedright\arraybackslash}p{0.18\textwidth}@{\hspace{8pt}}>{\raggedright\arraybackslash}p{\dimexpr\textwidth-0.40\textwidth-16pt\relax}@{}}
    \toprule
    Component & Model & Purpose / output constraints \\
    \midrule
    Initial generation & $M_{\mathrm{seed}}$ & Produce one Lean~4 file with a single theorem statement (\texttt{:= by sorry}). \\
    Operators (full/diff/cross) & $M_{\mathrm{patch}}$ & Edit-conditioned rewrite of a parent; cross additionally conditions on one inspiration from the archive. \\
    Compile repair (bounded) & $M_{\mathrm{patch}}$ & Minimal patch conditioned on compiler feedback; debited within the same call budget. \\
    Semantic repair (bounded) & $M_{\mathrm{patch}}$ & Revision conditioned on the informal statement and judge feedback; debited within the budget. \\
    Semantic judge & \path{CriticLean-Qwen3-14B} & Binary verdict on statement faithfulness (proof ignored), returned as JSON (\path{Correct/Incorrect} plus a short rationale). \\
    Prover & \path{Goedel-Prover-V2-32B} & Complete a Lean file by replacing the placeholder proof under a fixed prover prompt. \\
    \bottomrule
  \end{tabular}
\end{table*}
\FloatBarrier

\subsection{Statement generation (initial)}
\label{sec:appendix-prompts-initial}
\paragraph{Prompt template.}
\begin{PromptBox}
You are an expert in Lean 4 theorem proving and Mathlib.

Given a problem statement in natural language, write a COMPLETE Lean 4 file (imports + theorem)
that formalizes the mathematical content.

MANDATORY OUTPUT REQUIREMENTS:
- Output EXACTLY ONE Lean 4 code block: ```lean ... ```.
- The code MUST start with the following two lines (in this order):
  import Mathlib
  import Aesop
- You MAY add additional `import ...` / `open ...` / `set_option ...` lines after that if needed (unless a mode-specific prefix locks the header, e.g., diff mode).
- Do NOT include any comments.
- Include EXACTLY ONE `theorem` declaration.
- The theorem MUST end with `:= by sorry` (do not provide a proof).
- Use Lean 4 v4.15-compatible syntax and Mathlib definitions.

Natural language statement:
{informal}

Return ONLY the Lean 4 code block.
\end{PromptBox}

\subsection{Evolution operators (full / diff / cross)}
\label{sec:appendix-prompts-evolution}
\paragraph{Shared prompt skeleton.}
\begin{PromptBox}
You are an expert in Lean 4 theorem proving and the Mathlib library.
Your task is to improve a Lean 4 formalization for a given natural language mathematical claim.

Hard requirements (must satisfy all):
- Output a COMPLETE Lean 4 file (imports + exactly one theorem) in a single ```lean``` code block.
- The code MUST start with these two lines (in this order):
  import Mathlib
  import Aesop
- You MAY add additional `import ...` / `open ...` / `set_option ...` lines after that if needed.
- Do NOT include any comments.
- Include EXACTLY ONE `theorem` declaration.
- The theorem MUST end with `:= by sorry` (do not provide a proof).

Quality goals:
- compile_ok: The file compiles without errors.
- semantic_ok: The theorem statement matches the informal mathematical meaning.

Natural language statement:
{informal}

Current Lean 4 program:
```lean
{code_content}
```

Evaluation results:
{performance_metrics}{text_feedback_section}

Return ONLY the Lean 4 code block.
\end{PromptBox}

\paragraph{Full-mode addenda (sampled variants; one per call).}
\begin{PromptBox}
(i) Default rewrite: revise types/hypotheses/structure as needed.
(ii) Different interpretation: try a different reasonable formalization of the same claim.
(iii) Type optimization: improve binder structure and type-class constraints.
(iv) Mathlib alignment: prefer standard Mathlib names and definitions.
(v) Simplification: if possible, state a simpler claim without changing meaning.
\end{PromptBox}

Diff mode prepends a short local-edit instruction (still returning a full file), while cross mode additionally provides one
sampled inspiration candidate from the archive/top-k pool as extra context.
\paragraph{Diff-mode prefix (local edit).}
\begin{PromptBox}
You are doing a LOCAL EDIT of a Lean 4 formalization file.
Header lock (diff mode only):
- Do NOT add/remove/reorder/modify any preamble lines (imports/opens/options).
- Keep everything before the `theorem` declaration EXACTLY unchanged.
Goal:
- Make the smallest possible change to improve compilation and formalization quality.
- Keep changes minimal; do NOT change the intended mathematical meaning.
\end{PromptBox}

\paragraph{Cross-mode prefix (inspiration-guided).}
\begin{PromptBox}
You are doing CROSS patching.
You are given one inspiration candidate from the archive/top-k pool as extra context.
Your task is to improve the current formalization by optionally borrowing useful structure from the inspiration.
\end{PromptBox}

\paragraph{Cross-mode extra context (one per call).}
\begin{PromptBox}
[Inspiration candidate]
```lean
{inspiration_code_1}
```
\end{PromptBox}

\subsection{Bounded repair prompts}
\label{sec:appendix-prompts-repair}
Repairs count toward the generator-call budget. Compilation repair triggers only when \texttt{compile\_ok}=0; semantic repair triggers only when \texttt{compile\_ok}=1 and \texttt{semantic\_ok}=0.

\paragraph{Compilation repair prompt.}
\begin{PromptBox}
You are an expert in Lean 4 theorem proving and Mathlib.
You are doing SYNTAX / COMPILATION REPAIR.

You will receive:
- A natural language statement (the semantic target)
- A Lean 4 file that fails to compile
- Compiler error feedback

Your job:
- Produce a corrected Lean 4 file that compiles.

IMPORTANT:
- Fix compilation only; do NOT change the intended mathematical meaning.
- Keep changes minimal (types, identifiers, imports, binder annotations, etc.).

Natural language statement:
{informal}

Current Lean 4 code (does NOT compile):
<CURRENT_CODE>
{original_code}
</CURRENT_CODE>

Compiler feedback:
- Error type: {compile_error_type}
- Error message:
```
{compile_error_msg}
```

Return exactly one Lean 4 code block, starting with `import Mathlib` and `import Aesop`,
with exactly one theorem ending in `:= by sorry`.
\end{PromptBox}

\clearpage
\paragraph{Semantic repair prompt.}
\begin{PromptBox}
You are an expert in mathematics and Lean 4 (Mathlib).
You are doing SEMANTIC REPAIR.

You will receive:
- A natural language statement (the semantic target)
- A Lean 4 file that is intended to formalize it
- Critic feedback (Accuracy Confirmation) describing mismatches

Your job:
- Modify the Lean 4 code so that the theorem statement matches the natural language statement.

Rules:
- You MAY change hypotheses and the conclusion if needed to match the semantics.
- You MAY adjust/add imports/opens/options as needed to keep the file compiling.
- Do NOT "solve" the task by weakening it to `True` or a tautology.
- Do NOT include any comments.
- Output a complete Lean 4 file starting with:
  import Mathlib
  import Aesop
- Include exactly one theorem, ending with `:= by sorry`.

Natural language statement:
{informal}

Current Lean 4 code:
<CURRENT_CODE>
{original_code}
</CURRENT_CODE>

Critic feedback (Accuracy Confirmation):
{critic_feedback}

Goal:
- Modify the Lean 4 theorem so that the semantic judge would accept it as an exact formalization of the natural-language statement.
- Address every mismatch mentioned in the Critic feedback.

Return exactly one Lean 4 code block, starting with `import Mathlib` and `import Aesop`,
with exactly one theorem ending in `:= by sorry`.
\end{PromptBox}

\subsection{Semantic judge prompt (CriticLean-Qwen3-14B)}
\label{sec:appendix-prompts-judge}
We use CriticLean-Qwen3-14B as the LLM-as-judge implementation of $\SemOK$, judging semantic consistency between the natural-language statement and the Lean 4 theorem statement (proof ignored).
The prompt template below is used for each semantic check.

\begin{PromptBox}
Role: Lean 4 & Formal Verification Expert
Input:
- Mathematical_Text: A math problem and its answer (no proof).
- Lean4Code: A Lean 4 theorem statement formalizing the problem (proof intentionally omitted).
Goal:
Determine if the Lean 4 theorem statement is an exact and faithful formalization of the mathematical problem.
Do not evaluate or consider the answer or the proof. Your sole task is to verify the correctness of the formalization.

Evaluation stages:
1. Identify the mathematical claim: variables, types, quantifiers, assumptions, constraints, logic structure, and conclusion.
2. Extract the corresponding components from the Lean 4 statement, ignoring the proof.
3. Compare the two statements for semantic alignment, quantifier correctness, preserved constraints/boundaries, correct typing, syntactic validity, and absence of semantic drift or unjustified additions.
4. Give the final judgment using only what is explicitly expressed in the Lean statement, not the proof.
5. If correct, explain why the elements match; if incorrect, list mismatches and their effect on correctness.

Output Format:
Return exactly one JSON object:
{
  "reasons": "<your detailed analysis as a single string>",
  "is_assistant_correct": "Correct or Incorrect"
}

Input Data:
-- Start of Mathematical_Text --
{informal}
-- End of Mathematical_Text --
-- Start of Lean4Code --
{lean_statement}
-- End of Lean4Code --
\end{PromptBox}

\subsection{Prover prompt (Goedel-Prover-V2-32B)}
\label{sec:appendix-prompts-prover}
\label{sec:appendix-prover-prompt}
We evaluate downstream proof utility using a fixed prover configuration (Goedel-Prover-V2-32B) and a fixed prompt template.
The prompt asks the prover to complete a Lean file by replacing the placeholder proof.
\begin{PromptBox}
Complete the following Lean 4 code.
Return ONLY a complete Lean 4 file inside a ```lean4``` code fence (no explanations): {lean_file}
\end{PromptBox}
\FloatBarrier

\FloatBarrier

\end{document}